\PassOptionsToPackage{table}{xcolor}
\documentclass[10pt]{article}
\usepackage{colm2024_conference}
\usepackage{graphicx}
\usepackage{float}

\usepackage{amsmath,amsfonts,bm}

\def\eqref#1{equation~\ref{#1}}

\def\plaineqref#1{\ref{#1}}

\def\1{\bm{1}}

\DeclareMathAlphabet{\mathsfit}{\encodingdefault}{\sfdefault}{m}{sl}
\SetMathAlphabet{\mathsfit}{bold}{\encodingdefault}{\sfdefault}{bx}{n}

\usepackage{url}
\usepackage{booktabs}
\usepackage{array}
\usepackage{makecell}
\usepackage{adjustbox}
\usepackage[skip=10pt]{caption}
\usepackage{multirow}
\usepackage{amssymb}
\usepackage[table]{xcolor}
\usepackage[most]{tcolorbox}
\usepackage{tikz}
\usetikzlibrary{positioning,calc,arrows.meta,fit,backgrounds}

\definecolor{cAttn}{HTML}{F3C9C9}
\definecolor{cMoE}{HTML}{CDE9D2}
\definecolor{cShared}{HTML}{BFE3C6}
\definecolor{cRouted}{HTML}{C9CFEC}
\definecolor{cGrey}{HTML}{EDEDED}
\definecolor{cRes}{HTML}{9C2B2B}

\newcolumntype{L}[1]{>{\raggedright\arraybackslash}p{#1}}


\definecolor{PromptBg}{HTML}{F7F7F9}
\definecolor{PromptFrame}{HTML}{C8CBD0}
\newtcolorbox{promptbox}{%
  enhanced, sharp corners,
  boxrule=0.4pt, colback=PromptBg, colframe=PromptFrame,
  left=7pt, right=7pt, top=6pt, bottom=6pt,
  before skip=2pt, after skip=0pt,
  before upper={\footnotesize\ttfamily\raggedright\setlength{\parindent}{0pt}}}

\usepackage{XCharter}
\definecolor{InstellaRowBlue}{HTML}{D2EFFC}
\definecolor{LinkGray}{RGB}{237,28,36}
\definecolor{TitleGray}{RGB}{0,0,0}
\definecolor{ContentGray}{RGB}{38,38,38}
\definecolor{AbstractBg}{HTML}{ffffff}
\definecolor{AbstractTopLine}{HTML}{9d9fa2}
\usepackage[colorlinks=true, pdfborder={0 0 0}, linkcolor=LinkGray, citecolor=LinkGray, urlcolor=LinkGray]{hyperref}

\renewenvironment{abstract}
  {%
   \noindent\fontsize{10}{13}\selectfont\ignorespaces}
  {%
   \begin{center}\fontsize{10}{12}\selectfont
     \huggingface\ \href{\hflink}{Models}\qquad
     \github\ \href{\ghlink}{Code}\qquad
     \blog\ \href{\bloglink}{Blog}
   \end{center}%
   \par\vspace{0.6em}}

\DeclareCaptionFont{cap}{\fontsize{10}{12}\selectfont}
\makeatletter
\renewcommand\section{\@startsection{section}{1}{\z@}{-2.0ex plus -0.5ex minus -.2ex}{1.5ex plus 0.3ex minus0.2ex}{\bfseries\color{TitleGray}\fontsize{14}{16}\selectfont\raggedright}}
\renewcommand\subsection{\@startsection{subsection}{2}{\z@}{-1.8ex plus -0.5ex minus -.2ex}{0.8ex plus .2ex}{\bfseries\color{TitleGray}\fontsize{13}{16}\selectfont\raggedright}}
\renewcommand\subsubsection{\@startsection{subsubsection}{3}{\z@}{-1.5ex plus -0.5ex minus -.2ex}{0.5ex plus .2ex}{\bfseries\color{TitleGray}\fontsize{11}{14}\selectfont\raggedright}}
\makeatother

\usepackage{tocloft}
\definecolor{TocSubGray}{RGB}{140,140,140}

\makeatletter
\let\origl@section\l@section
\renewcommand*\l@section[2]{\begingroup\hypersetup{linkcolor=ContentGray}\origl@section{#1}{#2}\endgroup}
\let\origl@subsection\l@subsection
\renewcommand*\l@subsection[2]{\begingroup\hypersetup{linkcolor=TocSubGray}\origl@subsection{#1}{#2}\endgroup}
\let\origl@subsubsection\l@subsubsection
\renewcommand*\l@subsubsection[2]{\begingroup\hypersetup{linkcolor=TocSubGray}\origl@subsubsection{#1}{#2}\endgroup}
\makeatother

\newcommand{\shorttitle}{Instella-MoE Technical Report}
\title{\bfseries\fontsize{26}{28}\selectfont \shorttitle}

\usepackage{fancyhdr}
\renewcommand{\headrulewidth}{1.5pt}

\fancypagestyle{firstpage}{%
  \fancyhf{}%
  \fancyhead[R]{\raisebox{-0.2cm}{\includegraphics[height=0.5cm]{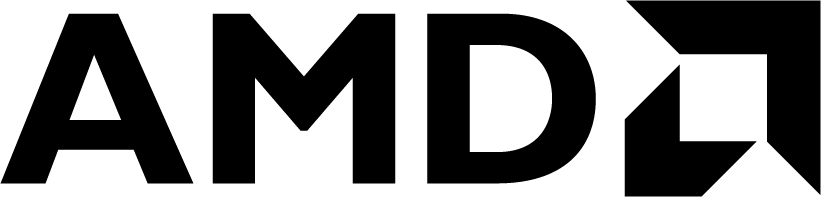}}}%
  \renewcommand{\headrulewidth}{1.2pt}%
  \renewcommand{\headrule}{\hbox to\headwidth{\color{black}\leaders\hrule height \headrulewidth\hfill}}%
}

\makeatletter
\def\@maketitle{\vbox{\hsize\textwidth
  \vspace*{0.65cm}%
  {\raggedright\bfseries\fontsize{24}{28}\selectfont\color{ContentGray} \@title\par}
  \vspace{0.6em}
  {\raggedright\@author\par}
  \vspace{0.3em}

  \vskip 0.02in
}\thispagestyle{firstpage}}
\makeatother

\newcommand{\coremark}{\textcolor{LinkGray}{\ensuremath{\blacktriangle}}}
\newcommand{\leadmark}{\textcolor{orange!100!black}{\ensuremath{\bigstar}}}
\author{%
{\bfseries \fontsize{10}{14.5}\selectfont Jiang Liu \textsuperscript{\leadmark\coremark}, Sudhanshu Ranjan \textsuperscript{\coremark}, Prakamya Mishra \textsuperscript{\coremark}, Yonatan Dukler \textsuperscript{\coremark}, Gowtham Ramesh \textsuperscript{\coremark}, Jialian Wu, Ximeng Sun, Wen Xie, Chaojun Hou, Vikram Appia, Zhenyu Gu, Zicheng Liu \textsuperscript{\coremark}, Emad Barsoum}\\[0.25em]
{\normalfont \textsuperscript{\leadmark}\it { \fontsize{10}{12}\selectfont Project lead}\quad \textsuperscript{\coremark}\it { \fontsize{10}{12}\selectfont Core contributors}}
}

\makeatletter
\@ifundefined{ps@firstpage}{\let\ps@firstpage\ps@fancy}{}
\@ifundefined{ps@undefinedpagestyle}{\let\ps@undefinedpagestyle\ps@plain}{}
\makeatother
\begin{document}

\maketitle
\begin{abstract}
\textbf{In this work, we introduce Instella-MoE, a fully open Mixture-of-Experts (MoE) language model with 16 billion total parameters and 2.8 billion active parameters per token, trained entirely from scratch on AMD Instinct\texttrademark{} MI300X and MI325X GPUs. Instella-MoE combines a sparsely activated MoE design with architectural and system-level innovations, including Gated Multi-head Latent Attention (Gated MLA) and FarSkip-Collective connectivity, enabling efficient large-scale training and inference. The model is developed through a multi-stage pipeline comprising pre-training, mid-training, long-context extension, supervised fine-tuning with feedback-driven data curation, direct preference optimization, and reinforcement learning with Multi-Teacher On-Policy Distillation. Instella-MoE achieves an average score of 76.7 across standard pre-training benchmarks, outperforming prior fully open models including OLMo-3-7B, SmolLM3-3B, and OLMoE-1B-7B, while remaining competitive with open-weight MoE and dense baselines at comparable active-parameter scales, including Moonlight-16B-A3B and Qwen3.5-4B. After post-training, our final Think checkpoint achieves an average score of 73.2 across instruction-following, reasoning, math, coding, and chat benchmarks, outperforming both fully open and open-weight models with comparable or larger active-parameter counts in our evaluation. To support transparent and reproducible research, we release the complete Instella-MoE model flow, including model weights, training configurations, data mixtures, and training code. Together, these contributions establish Instella-MoE a strong, fully open foundation for efficient, high-performing MoE models and reproducible research.}
\end{abstract}

\begin{figure}[H]
    \centering
    \includegraphics[width=\linewidth]{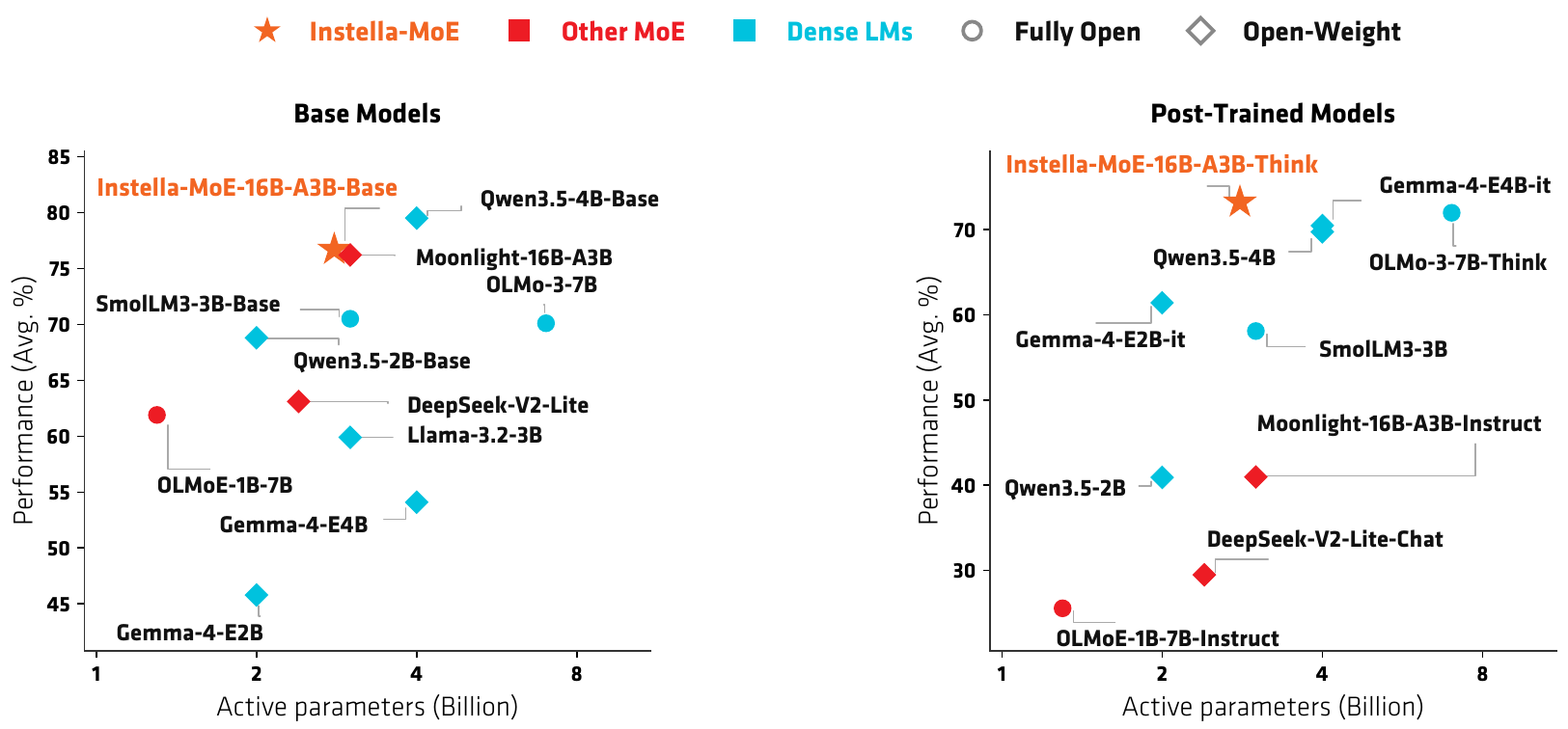}
    \caption{\textbf{Performance vs.\ active parameters.} \emph{Left:} base models. \emph{Right:} post-trained models. Color distinguishes MoE and dense architectures; marker shape distinguishes fully open and open-weight models.}
    \label{fig:cost_performance}
\end{figure}

\vspace{0.5em}
{\tableofcontents}
\vspace{1.0em}

\newpage
\section{Introduction}
The rapid advancement of artificial intelligence (AI), driven in large part by large language models (LLMs)~\citep{gemini31,gpt56,claudefable5}, has accelerated progress toward artificial general intelligence and transformed society at large. However, much of this progress continues to be led by proprietary releases (e.g., GPT-5.6~\citep{gpt56}, Claude Fable~5~\citep{claudefable5}, Gemini~3.1 Pro~\citep{gemini31}, and Muse Spark~\citep{musespark}), where model weights and architecture are not released and training data, methods, and evaluation protocols remain opaque. While these models have set new state-of-the-art performance, their closed nature hinders scientific understanding, reproducibility, and equitable access.

In response, leading AI labs have increasingly released open-weight models whose trained parameters are available under permissive licenses. Recent systems such as GLM-5.3~\citep{glm53}, Qwen3.8-Max~\citep{qwen38max}, Kimi~K3~\citep{kimik3}, Gemma~4~\citep{gemma4}, DeepSeek-V4~\citep{deepseekv4}, Llama~4~\citep{llama4}, and Mistral Large~3~\citep{mistral3} demonstrate that high-performing models are increasingly available outside proprietary labs. Yet most of these releases remain open-weight rather than fully open: their pre-training corpora, preprocessing pipelines, and complete training recipes are either undisclosed or only partially documented. As a result, researchers cannot fully reproduce results, audit potential data contamination, or study how architectural and data choices interact at scale.

At the same time, sparsely activated Mixture-of-Experts (MoE) architectures have become a central strategy for improving the cost--performance trade-off of large language models~\citep{olmoe,llama4,deepseekv4,qwen38max}. By activating only a small subset of parameters per token, MoE models can achieve the representational capacity of much larger dense models while keeping per-token compute closer to that of a smaller dense baseline. This design now appears across leading open-weight systems~\citep{qwen38max, deepseekv4, kimik3, glm53}. Despite these advances, fully open MoE models with competitive performance and complete training transparency remain comparatively rare. OLMoE~\citep{olmoe} established an important baseline with 6.9B total parameters and 1.3B active parameters, while Marco-MoE~\citep{jiang2026marcomoeopenmultilingualmixtureofexpert} extended fully open sparse training to multilingual settings through dense-checkpoint upcycling.

In this work, we introduce Instella-MoE, a fully open Mixture-of-Experts language model with 16 billion total parameters and 2.8 billion active parameters per token, trained entirely from scratch on AMD Instinct\texttrademark{} MI300X and MI325X GPUs. Instella-MoE combines a decoder-only MoE architecture with two key innovations: Gated Multi-head Latent Attention (Gated MLA), which augments Multi-head Latent Attention (MLA)~\citep{deepseekv2} with a lightweight gating mechanism~\citep{gatedattention} to improve model expressivity, and FarSkip-Collective~\citep{dukler2026farskipcollectiveunhobblingblockingcommunication}, an MoE connectivity pattern that overlaps expert-parallel communication with computation to improve the efficiency of both training and inference.

We train Instella-MoE using a multi-stage pipeline spanning pre-training, mid-training, long-context extension, supervised fine-tuning, direct preference optimization, and reinforcement learning. Pre-training uses 7.1 trillion tokens from high-quality open corpora, including web text, mathematics, and code. During mid-training, we train on curated STEM- and reasoning-focused mixtures and combine checkpoints from multiple data-mixture variants through model souping. We then extend the context length from 4K to 64K tokens using general and domain-targeted long-context mixtures, together with YaRN RoPE scaling~\citep{peng2024yarn} and document masking. Starting from the resulting base checkpoint, post-training proceeds in three stages: (i) supervised fine-tuning (SFT) with feedback-driven data curation targeting model weaknesses; (ii) direct preference optimization (DPO)~\citep{dpo} on contrastive preference data, with the auxiliary load-balancing loss and router bias updates disabled to reduce expert-routing drift; and (iii) reinforcement learning (RL) focused on instruction following, followed by Multi-Teacher On-Policy Distillation (MOPD)~\citep{ma2026mopd}. The entire training pipeline is implemented on AMD hardware using the open-source \textbf{Primus}~\citep{primus} and \textbf{Miles}~\citep{radixark2025miles} frameworks.

With only 2.8B active parameters per token, the Instella-MoE base checkpoint achieves an average score of \textbf{76.7} across standard pre-training benchmarks. It outperforms prior fully open models, including OLMo-3-7B~\citep{olmo2025olmo}, SmolLM3-3B~\citep{bakouch2025smollm3}, and OLMoE-1B-7B~\citep{olmoe}, while remaining competitive with open-weight MoE and dense baselines with comparable or larger active-parameter counts, including Moonlight-16B-A3B~\citep{liu2025muonscalablellmtraining}, Gemma-4-E4B~\citep{gemma4}, and Qwen3.5-4B~\citep{qwen3.5}. After the full post-training pipeline, our final Think checkpoint achieves an average score of \textbf{73.2} across instruction-following, reasoning, math, coding, and chat benchmarks, outperforming the fully open OLMo-3-7B-Think (72.0) as well as the open-weight models Gemma-4-E4B (70.5) and Qwen3.5-4B (69.7), and yielding the strongest overall post-trained result in our comparison. These quality gains are complemented by architectural and system-level efficiency improvements, with FarSkip-Collective increasing pre-training throughput by 12.7\% through overlapping expert-parallel communication, while expert-parallel serving with SGLang raises time-to-first-token throughput by up to 39.2\%.

In line with prior fully open model releases~\citep{olmo,olmo2,olmo2025olmo,allal2024smollm,allal2025smollm2smolgoesbig,bakouch2025smollm3,liu2025instella,wang2025instellat2i,sun2025instellavl}, we release the \textbf{complete Instella-MoE model flow}: checkpoints from every major training stage---pre-training, mid-training, long-context extension, SFT, DPO, and RL---together with training configurations (Table~\ref{tab:training_hparams}), data mixtures (Table~\ref{tab:training_data}), training code, and evaluation protocols. The complete release enables the community to reproduce, audit, and extend every stage of MoE model development on AMD platforms, supporting transparent benchmarking and further research into fully open language models.

To summarize, our contributions are fourfold:
\begin{itemize}
    \item \textbf{Instella-MoE}, a fully open MoE model with 16B total parameters and 2.8B active parameters per token, trained entirely from scratch on AMD Instinct GPUs, featuring Gated MLA for enhanced attention expressivity and FarSkip-Collective for efficient model training and inference.
    \item \textbf{A complete multi-stage training pipeline}, spanning pre-training, mid-training, long-context extension, SFT with feedback-driven data curation, DPO, and instruction-following RL followed by MOPD.
    \item \textbf{State-of-the-art fully open performance}, with the base checkpoint averaging 76.7 on pre-training benchmarks and the final Think checkpoint averaging 73.2 on post-training benchmarks, both matching or outperforming fully open and open-weight baselines at comparable active-parameter scales.
    \item \textbf{Full transparency and reproducibility}, through the release of stage-by-stage model weights, training recipes, data mixtures, training code, and evaluation protocols.
\end{itemize}

\section{Instella-MoE}

\subsection{Model Architecture}
\subsubsection{Architecture Overview}
\begin{figure}[t]
    \centering
    \resizebox{\linewidth}{!}{\input{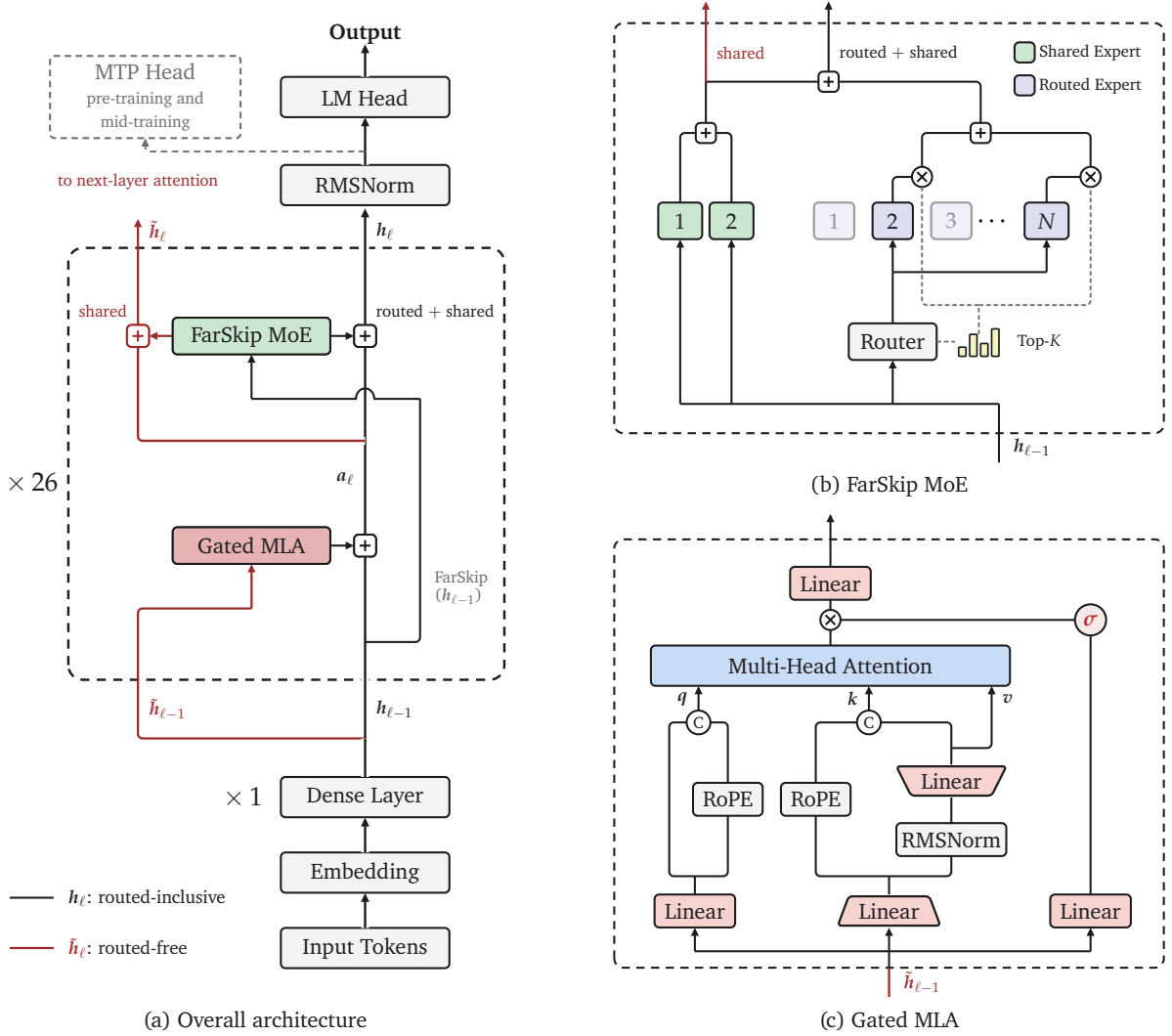}}
    \caption{\textbf{Instella-MoE-16B-A3B architecture.} \emph{(a)} Overall architecture. FarSkip-Collective allows the Dispatch and Combine all-to-all collectives to overlap with attention and shared-expert computation. The MTP head is attached during pre-training and mid-training. \emph{(b)} FarSkip-Collective
    MoE block. The MoE block returns both the shared and routed+shared experts explicitly. \emph{(c)} Simplified Gated MLA block. An input-conditioned sigmoid gate
    modulates the per-head attention outputs before the output projection.}
    \label{fig:architecture}
\end{figure}

Instella-MoE-16B-A3B is a text-only, decoder-only Transformer~\citep{attention} with 27 layers, of which 26 are sparsely activated Mixture-of-Experts (MoE) layers~\citep{shazeer2017outrageouslylargeneuralnetworks}, giving a total of 16B parameters and 2.8B active parameters per token. The MoE layers follow the standard shared-plus-routed expert design and dispatch each token to 6 of 64 routed experts alongside 2 shared experts. During pre-training and mid-training, we add a one-layer Multi-Token Prediction (MTP) head to support an auxiliary training objective. Instella-MoE extends the standard sparse-MoE architecture~\citep{deepseekv2,deepseekai2024deepseekv3technicalreport,liu2025muonscalablellmtraining} with two innovations: \emph{Gated Multi-head Latent Attention} (Gated MLA) to improve model quality and \emph{FarSkip-Collective} to improve compute efficiency. Gated MLA augments MLA~\citep{deepseekv2} with an input-conditioned output gate~\citep{gatedattention} that modulates individual attention-output channels (Sec.~\ref{sec:gated_mla}). FarSkip-Collective~\citep{dukler2026farskipcollectiveunhobblingblockingcommunication} modifies the layer connectivity to let attention and expert-parallel communication overlap with independent computation in the layer (Sec.~\ref{sec:farskip}). Figure~\ref{fig:architecture} provides an overview of the architecture, and Table~\ref{tab:model_arch} lists the model hyperparameters.

\paragraph{Mixture-of-Experts Layers.}
Instella-MoE follows the fine-grained, shared-plus-routed design of DeepSeek-V3~\citep{deepseekai2024deepseekv3technicalreport}, with the first decoder layer retained as a dense feed-forward network. Each of the remaining 26 MoE layers dispatches every token to the top $K{=}6$ of $N{=}64$ routed experts of intermediate size 1,408, alongside two shared experts implemented as a single fused feed-forward network of intermediate size 2,816. The router weights of the selected experts are scaled by 2.5 following DeepSeek-V3.

\paragraph{Load balancing.} Our primary expert load-balance mechanism follows the bias-based loss-free balancing in~\citep{deepseekai2024deepseekv3technicalreport}, where per-expert bias is added to the router score used for top-$K$ selection but not to the mixture weights. The bias is updated between steps toward the target load at a rate of $1\times10^{-3}$. This adjusts the top-$K$ assignments without introducing gradients from an auxiliary objective. We also use a sequence-level auxiliary load-balancing loss with coefficient $1\times10^{-4}$ as a secondary regularizer. During direct preference optimization, we freeze the bias updates and set the auxiliary-loss coefficient to zero to limit routing drift (Sec.~\ref{sec:dpo}).

\paragraph{Multi-Token Prediction.}
Following prior work~\citep{gloeckle2024betterfasterlarge,deepseekai2024deepseekv3technicalreport}, we attach a single Multi-Token Prediction (MTP) module during pre-training and mid-training. In addition to the standard next-token objective, the module predicts one additional future token at each position, providing a richer training signal. We weight the MTP loss by 0.3 during pre-training and reduce it to 0.1 during mid-training. The module is disabled during long-context extension and is not used at inference.

\subsubsection{Gated Multi-head Latent Attention}
\label{sec:gated_mla}
Multi-head Latent Attention (MLA)~\citep{deepseekv2} compresses each token's key--value states into a low-rank latent representation to reduce the KV-cache footprint. Standard MLA passes the concatenated per-head outputs of scaled dot-product attention directly through an output projection. Gated MLA retains the low-rank key--value factorization but inserts an input-conditioned gate between the attention output and the output projection (Figure~\ref{fig:architecture}(c)).

Following the best-performing configuration of \citet{gatedattention}, we use a multiplicative, sigmoid-activated element-wise gate. Let $\bm{x}_t\in\mathbb{R}^{d}$ be the RMSNorm-normalized layer input and $\hat{\bm{o}}_t\in\mathbb{R}^{H d_v}$ the concatenated output of scaled dot-product attention, with $H$ heads of value dimension $d_v$. The Gated MLA output is
\begin{equation}
    \bm{y}_t = \mathbf{W}_o\big[\operatorname{Sigmoid}(\mathbf{W}_g\bm{x}_t)\odot\hat{\bm{o}}_t\big].
    \label{eq:gated_mla}
\end{equation}
Here, $\mathbf{W}_g\in\mathbb{R}^{H d_v\times d}$ is the gate projection and $\mathbf{W}_o$ is the attention output projection. The gate allows each token to modulate individual value channels within each head. In our controlled 200B-token ablation, Gated MLA improves the average score from 49.86 to 50.33 over vanilla MLA, with the largest gains on MMLU and code generation (Table~\ref{tab:arch_ablation}).

\begin{table}[t]
    \centering
    \small
    \caption{\textbf{Architecture hyperparameters of Instella-MoE-16B-A3B.}}
    \begin{tabular}{ll}
    \toprule
    \textbf{Parameter} & \textbf{Value} \\
    \midrule
    \multicolumn{2}{@{}l}{\emph{Backbone}} \\
    Total / active parameters & 16B / 2.8B \\
    Decoder layers & 27 (1 dense $+$ 26 MoE) \\
    Hidden size & 2,048 \\
    Tokenizer / vocabulary size & DeepSeek-V3 / 128,896 \\
    Context length (base / extended) & 4,096 / 65,536 \\
    \midrule
    \multicolumn{2}{@{}l}{\emph{Attention}} \\
    Attention type & Gated MLA \\
    Attention heads & 16 \\
    KV latent rank & 512 \\
    QK dimensions (content / RoPE) & 96 / 32 \\
    Value-head dimension & 128 \\
    \midrule
    \multicolumn{2}{@{}l}{\emph{Mixture-of-Experts}} \\
    Routed experts $N$ / active per token $K$ & 64 / 6 \\
    Shared experts & 2 \\
    Routed expert FFN size & 1,408 \\
    Fused shared-expert FFN size & 2,816 \\
    Router top-$K$ scaling factor & 2.5 \\
    Connectivity & FarSkip-Collective \\
    \midrule
    \multicolumn{2}{@{}l}{\emph{Other}} \\
    Normalization & RMSNorm \\
    Activation & SwiGLU \\
    Positional embedding & RoPE \\
    Auxiliary objective & MTP (1 layer; pre-training and mid-training) \\
    \bottomrule
    \end{tabular}
    \label{tab:model_arch}
\end{table}

\subsubsection{FarSkip-Collective Connectivity}
\label{sec:farskip}
Expert parallelism (EP) is crucial for sparse MoE deployment. In Instella-MoE, we address its communication overhead through aggressive computation--communication overlap.
Concretely, the MoE layer is implemented in four main stages: router score computation, \emph{Dispatch}, expert computation, and \emph{Combine}.
The router stage computes the expert routing map. Next the Dispatch stage uses the map to send the relevant tokens to the corresponding GPUs via an all-to-all communication call.
After dispatching finishes, the GPUs compute the expert activations for their corresponding tokens. Lastly, the computed expert activations are mapped back to the original GPU placement via Combine, which runs an additional all-to-all communication call.
With this sequence of operations, Dispatch relies on the previous layer and the Combine call depends on the expert computation to finish. This set of MoE layer dependencies makes it so that neither all-to-all communication call can be overlapped with computation in the standard MoE architecture implementation.
To resolve this, with FarSkip-Collective we modify the transformer architecture to remove these dependencies and therefore enable overlapping of the Dispatch and Combine communication calls with computation that is not dependent on the communication output. In particular, with FarSkip-Collective the entire attention computation and shared expert computations become non-dependent and can therefore be overlapped with the Dispatch and Combine calls. We review the standard MoE and FarSkip-Collective architectures below.

Denote the output activation of a network after $k$ layers by $h_k$, and the $i$th sub-block (layer) of a standard Transformer by $f_i$. The standard Transformer output $h_k$ is computed as
\begin{equation}
    h_k = f_1(h_0) + f_2(h_1) + \dots + f_k(h_{k-1}).\label{eq:reg}
\end{equation}
Here the $f_i$ alternate between MLP / MoE blocks and Attention blocks and include normalization as part of the computation.
For large models, producing $f_i$ can involve blocking communication which stops $h_k$ from being used as an input to $f_{k+1}$ until $h_k$ is communicated between the GPUs -- leading to idle computation resources.
We propose to use an available activation instead, denoted as $\hat{h}_k$, to be used as input to $f_{k+1}$ and compute the next layer while the communication collective is producing $h_{k}$.
With this approach, $h_{k+1}$ is updated with $h_{k}$ once it is ready; however, the communication of $h_k$ can be overlapped over the duration of the computation of $f_{k+1}(\hat{h}_k)$:
\begin{equation}
\begin{aligned}
h_{k+1} &= h_{k} + f_{k+1}(\hat{h}_{k}) \\
    &= f_1(h_0) + f_2(\hat{h}_1) + \dots + f_{k+1}(\hat{h}_k)
\end{aligned}\label{eq:farskip}
\end{equation}
Like the standard transformer in Eq.~(\plaineqref{eq:reg}), all computed activations appear in the final hidden states in Eq.~(\plaineqref{eq:farskip}).
Comparing $\hat{h}_k$ with $h_k$, one can consider the partial or outdated activation $\hat{h}_k$ as an activation produced by omitting summands of the full activation sum. To maintain model capabilities while improving hardware performance, FarSkip-Collective is designed to allow for full communication overlapping arising from Dispatch, Combine, and the Attention block while incurring the most minimal activation omission in $\hat{h}_k$ as compared with $h_k$.
Concretely, to allow for Dispatch to overlap with the attention block, the input to the Dispatch call must not depend on the attention output, so we set
\begin{equation*}
    \text{moe-in}_k :=\hat{h}_k(\text{moe}) = h_{k-1}.
\end{equation*} This allows Dispatch and routing to initiate as soon as the previous layer's Combine finishes producing the tokens, and has the property that propagation is delayed by at most one layer. For the attention input, only the routed experts need to be communicated after expert computation, and they are therefore the only terms omitted from the attention input
\begin{equation*}
    \text{attn-in}_k := \hat{h}_k (\text{attn}) = h_{k-2} + \text{attn-out}_{k-1} + \text{shared-exp-out}_{k-1}.
\end{equation*}
This input to attention allows for overlapping the routed expert Combine with both shared expert computation and the first part of attention computation. Lastly, any Attention block communication is overlapped during routed-expert computation.
Put together, without any pipelining, FarSkip-Collective increases the overlap window of Dispatch and Combine to be the full layer duration, with the exception of the router and routed-expert computations.

For Instella-MoE, we run large-scale architectural ablations to evaluate the capabilities of Instella-MoE with the FarSkip-Collective architecture and observe comparable performance relative to standard MoE connectivity.
In particular, we train Instella-MoE (+FarSkip) for 200B pre-training tokens and ablate the architecture while fixing all training settings and seeds against Instella-MoE without FarSkip-Collective. In Table~\ref{tab:arch_ablation} we observe that the Instella-MoE (+FarSkip) variant achieves 50.38 average performance compared to 50.33 average performance when ablating the FarSkip-Collective connectivity with the standard transformer. In Appendix~\ref{app:overlapped_implementation} we describe the implementation techniques that enable communication-overlapped training and inference of Instella-MoE on AMD GPUs.

\subsection{Training Pipeline Overview}
\label{sec:pipeline}
\begin{figure}[t]
    \centering
    \includegraphics[width=\linewidth]{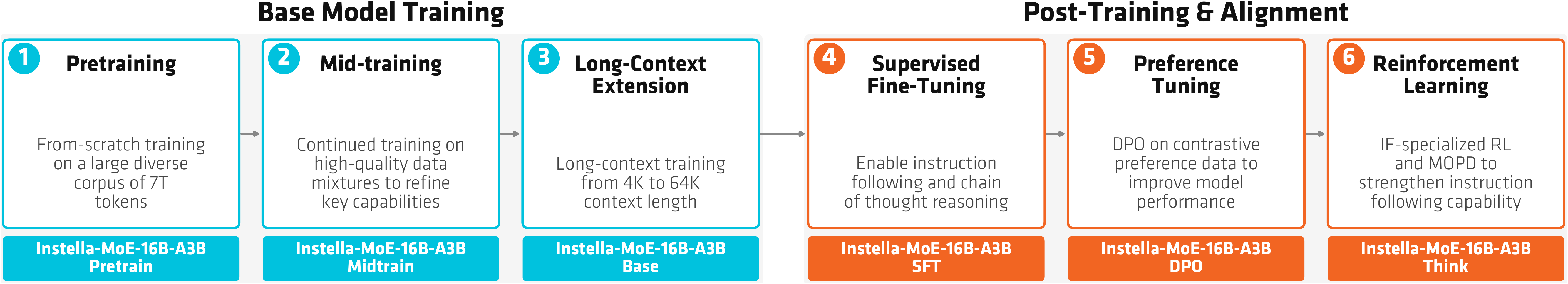}
    \caption{\textbf{Instella-MoE training pipeline.} Base-model training spans pre-training, mid-training, and long-context extension; post-training and alignment span SFT, preference tuning (DPO), and reinforcement learning.}
    \label{fig:training_pipeline}
\end{figure}

Instella-MoE is trained from scratch through the multi-stage pipeline shown in Figure~\ref{fig:training_pipeline}, which progressively builds capability from broad language understanding to long-context processing, instruction following, alignment, and reasoning. Base-model training comprises 7.1T-token pre-training at 4K context length (Sec.~\ref{sec:pretrain}), ${\sim}$100B-token mid-training on STEM- and reasoning-focused mixtures combined by model souping (Sec.~\ref{sec:midtrain}), and a two-stage extension of the context window from 4K to 64K that produces the Instella-MoE-16B-A3B-Base model (Sec.~\ref{sec:longcontext}). Post-training initializes from that base model and applies supervised fine-tuning with feedback-driven data curation (Sec.~\ref{sec:sft}), direct preference optimization with delta learning (Sec.~\ref{sec:dpo}), and instruction-following RL followed by multi-teacher on-policy distillation (Sec.~\ref{sec:rl}), producing Instella-MoE-16B-A3B-Think. We release the checkpoint of every training stage (Table~\ref{tab:checkpoints}) together with the per-stage hyperparameters (Table~\ref{tab:training_hparams}) and data mixtures (Table~\ref{tab:training_data}).

\subsection{Training Infrastructure}

All stages up to and including DPO run in the open-source \textbf{Primus} framework~\citep{primus} on its Megatron-LM~\citep{shoeybi2019megatron} backend using ROCm on AMD Instinct MI300X and MI325X GPUs. These stages use expert parallelism (EP${=}8$) over the 64 routed experts, tensor parallelism, FlashAttention~\citep{dao2023flashattention2}, and bfloat16 mixed precision; they also overlap gradient and parameter communication with computation. Pre-training and mid-training additionally overlap the expert-parallel all-to-all collectives using the FarSkip-Collective implementation described in Appendix~\ref{app:overlapped_implementation}.

Reinforcement learning is performed with \textbf{Miles}~\citep{radixark2025miles, slime_github}, an open-source PyTorch-native stack for large-scale RL post-training built on the slime ecosystem. Miles pairs a Megatron-LM learner with a pool of SGLang~\citep{zheng2024sglang} rollout engines under Ray-based multi-node orchestration and supports both asynchronous off-policy and synchronous on-policy rollout loops. Our two RL stages use these modes, respectively. We run Miles with Primus as the policy-training backend and SGLang as the rollout backend; Sec.~\ref{sec:rl} details the resulting actor--learner configuration.

\begin{table}[t]
\centering
\small
\caption{\textbf{Released Instella-MoE checkpoints.} We open-source weights from every major training stage to support reproducible model research.}
\setlength{\tabcolsep}{4pt}
\begin{adjustbox}{width=\linewidth,center}
\begin{tabular}{lll}
\toprule
\textbf{Checkpoint} & \textbf{Stage} & \textbf{Description} \\
\midrule
\href{https://huggingface.co/amd/Instella-MoE-16B-A3B-Pretrain}{\huggingface}\ Instella-MoE-16B-A3B-Pretrain & Pre-training & From-scratch MoE checkpoint after 7.1T-token pre-training. \\
\href{https://huggingface.co/amd/Instella-MoE-16B-A3B-Midtrain}{\huggingface}\ Instella-MoE-16B-A3B-Midtrain & Mid-training & STEM- and reasoning-sharpened checkpoint produced through model souping. \\
\href{https://huggingface.co/amd/Instella-MoE-16B-A3B-Base}{\huggingface}\ Instella-MoE-16B-A3B-Base & Long-context & Final base model after 64K long-context extension. \\
\href{https://huggingface.co/amd/Instella-MoE-16B-A3B-SFT}{\huggingface}\ Instella-MoE-16B-A3B-SFT & SFT & Instruction-tuned checkpoint with chain-of-thought capability. \\
\href{https://huggingface.co/amd/Instella-MoE-16B-A3B-DPO}{\huggingface}\ Instella-MoE-16B-A3B-DPO & DPO & Preference-tuned checkpoint with delta learning. \\
\href{https://huggingface.co/amd/Instella-MoE-16B-A3B-Think}{\huggingface}\ Instella-MoE-16B-A3B-Think & RL & Final Think checkpoint after IF-RL and MOPD. \\
\bottomrule
\end{tabular}
\end{adjustbox}
\label{tab:checkpoints}
\end{table}

\begin{table}[t]
\centering
\small
\caption{\textbf{Training hyperparameters by stage.} All stages use expert parallelism (EP${=}8$), bfloat16 mixed precision, and the Primus and Miles training stack unless otherwise noted.}
\begin{adjustbox}{width=\linewidth,center}
\setlength{\tabcolsep}{3.5pt}
\begin{tabular}{@{}lcccccc@{}}
\toprule
\textbf{Stage} & \textbf{Tokens / Data} & \textbf{Seq.\ Len.} & \textbf{Global BS} & \textbf{Peak LR} & \textbf{LR Schedule} & \textbf{Key Settings} \\
\midrule
Pre-training & 7.1T & 4,096 & 4,096 & $4\times10^{-4}$ & WSD & MTP, EOD masking, FarSkip \\
Mid-training & $\sim$100B & 4,096 & 1,024 & $2\times10^{-4}$ & Linear & 3 mixture variants + souping \\
Long-ctx Stage 1 & $\sim$194B & 65,536 & 128 & $2\times10^{-4}$ & Linear & YaRN ($\theta{=}8$M), doc masking \\
Long-ctx Stage 2 & $\sim$20B & 65,536 & 128 & $1\times10^{-4}$ & Linear & STEM-targeted data mix \\
SFT & $\sim$2.9M ex. & 32,768 & 128 & $1\times10^{-4}$ & Linear & 2 epochs, seq packing, feedback-driven curation \\
DPO & 150K pairs & 16,384 & 256 & $8\times10^{-8}$ & Linear & $\beta{=}5.0$, 0.75 epoch \\
IF-RL & 29.8K prompts & 16,384 rollout & 512 & $1\times10^{-6}$ & Constant & GRPO + DAPO + R3, 1,400 steps \\
MOPD & On-policy & 16,384 rollout & 512 & $1\times10^{-6}$ & Constant & Domain-routed teachers \\
\bottomrule
\end{tabular}
\end{adjustbox}
\label{tab:training_hparams}
\end{table}

\begin{table}[t]
\centering
\small
\caption{\textbf{Training data by stage.} All pre-training and mid-training corpora are drawn from publicly documented open datasets.}
\begin{adjustbox}{width=\linewidth,center}
\setlength{\tabcolsep}{4pt}
\begin{tabular}{@{}p{0.13\linewidth}p{0.69\linewidth}p{0.14\linewidth}@{}}
\toprule
\textbf{Stage} & \textbf{Primary Data Sources} & \textbf{License} \\
\midrule
Pre-training & Nemotron-CC-v2; Nemotron-Pretraining-SFT-v1; Nemotron-CC-Math-v1; MegaMath; FineMath; RefineCode; Nemotron-Pretraining-Code-v1; TxT360 & Source-dependent \\
\addlinespace[4pt]
Mid-training & Dolma~3 Dolmino 100B (web, code, math, science, instruction, reasoning) & ODC-BY-1.0 \\
\addlinespace[4pt]
Long-context & Dolma~3 Longmino 100B (Stage 1); Dolma~3 Dolmino 100B mix and full pool math/code/reasoning subsets; Instella-GSM8K-synthetic (Stage 2) & Source-dependent \\
\addlinespace[4pt]
SFT & Dolci-Think-SFT-7B; Nemotron-Cascade-2 (math, science); Nemotron-SFT-Competitive-Programming-v2 & Source-dependent \\
\addlinespace[4pt]
DPO & Dolci-Think-DPO-7B preference pairs & Source-dependent \\
\addlinespace[4pt]
RL & Dolci-Think-RL-7B (IF RLVR Mixture) & Source-dependent \\
\bottomrule
\end{tabular}
\end{adjustbox}
\label{tab:training_data}
\end{table}

\subsection{Pre-training}
\label{sec:pretrain}
\begin{figure}[t]
    \centering
    \includegraphics[width=\linewidth]{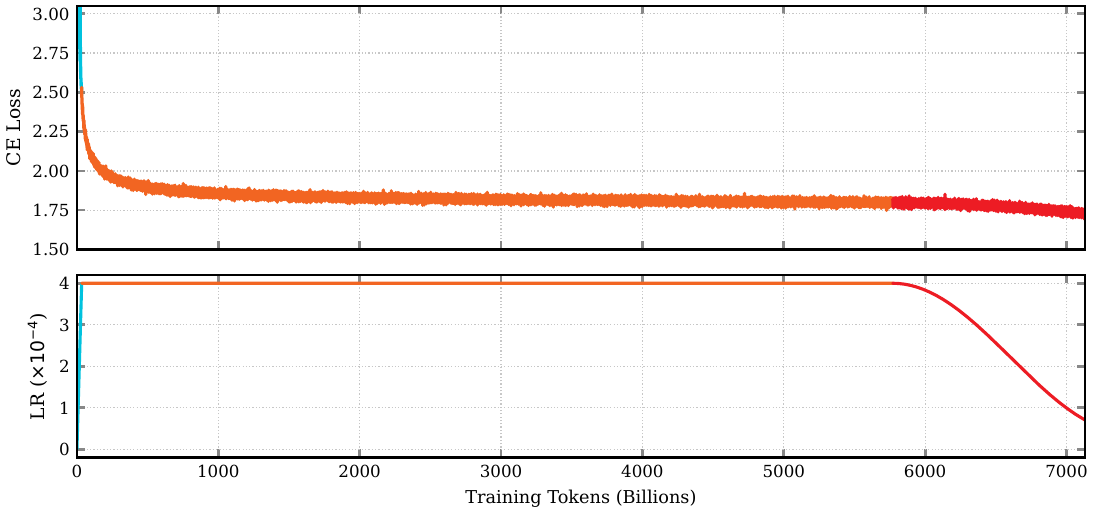}
    \caption{\textbf{Pre-training dynamics of Instella-MoE-16B-A3B over 7.1T tokens.} \emph{Top:} training cross-entropy (CE) loss. \emph{Bottom:} WSD learning-rate schedule with a peak of $4\times10^{-4}$.}
    \label{fig:pretrain_curves}
\end{figure}

We train Instella-MoE from scratch on 7.1 trillion tokens using a context length of 4,096 and a global batch size of 4,096. We use the AdamW optimizer with $\beta_1=0.9$, $\beta_2=0.95$, a weight decay of 0.1, and a WSD learning-rate scheduler with a peak learning rate of $4\times10^{-4}$, a minimum learning rate of $4\times10^{-5}$, and 2,000 warmup iterations. Figure~\ref{fig:pretrain_curves} shows the pre-training dynamics: after a short warmup, the learning rate stays at its $4\times10^{-4}$ peak for the bulk of training before decaying in the final phase, while the cross-entropy loss drops sharply early on and then decreases steadily to roughly $1.7$ by the end of the 7.1T-token run.

The pre-training mixture emphasizes high-quality open corpora spanning web text, mathematics, code, and science. We utilize Nemotron-CC-v2~\citep{su2025nemotroncc,nemotronnano2} for web data; Nemotron-CC-Math-v1~\citep{mahabadi2025nemotronccmath}, MegaMath~\citep{zhou2025megamath}, and FineMath~\citep{lozhkov2024finemath} for mathematics; RefineCode~\citep{huang2025opencoder} and Nemotron-Pretraining-Code-v1~\citep{nemotronnano2} for code; Nemotron-Pretraining-SFT-v1~\citep{nemotronnano2} for curated pre-training SFT-style data; and TxT360~\citep{txt360} for other domains. We tokenize all corpora with the DeepSeek-V3 tokenizer~\citep{deepseekai2024deepseekv3technicalreport}. During this stage, the MTP objective is active.

\subsection{Mid-training}
\label{sec:midtrain}
After pre-training, we perform mid-training on Dolma~3 Dolmino 100B~\citep{olmo2025olmo} to sharpen math, code, STEM, and reasoning capabilities. We train three mid-training variants (v1, v2, v3) that share the same Dolma~3 Dolmino 100B base mixture and differ only in a few STEM/reasoning subsets: \emph{v1} uses the original Dolmino 100B mix; \emph{v2} replaces the \texttt{STEM-Heavy Crawl} subset with its full version from the Dolma~3 Dolmino pool; and \emph{v3} additionally pulls in the full-pool subsets for \texttt{MegaMatt}, \texttt{General Reasoning Mix}, \texttt{Math Meta-Reasoning}, and \texttt{Code Meta-Reasoning}, expanding the mixture from $\sim$100B to $\sim$104B tokens (Table~\ref{tab:midtrain_variants}). Each variant is trained with a peak learning rate of $2\times10^{-4}$, a global batch size of 1,024, and linear learning-rate decay. The three resulting checkpoints are combined via equal-weight model souping to produce Instella-MoE-16B-A3B-Midtrain, which outperforms each individual variant (Table~\ref{tab:midtraining_model_souping}). The MTP objective remains active during this stage.

\begin{table*}[t]
\centering
\setlength{\tabcolsep}{3pt}
\renewcommand{\arraystretch}{1.1}
\scriptsize
\caption{\textbf{Mid-training model souping.} Comparison of the Instella-MoE-16B-A3B mid-training variants and their equal-weight model soup. HE+ denote HumanEval+.}
\resizebox{\textwidth}{!}{%
\begin{tabular}{l|*{13}{c}}
\toprule
\textbf{Model} & \textbf{ARC-E} & \textbf{ARC-C} & \textbf{BoolQ} & \textbf{SciQ} & \textbf{PIQA} & \textbf{HSwag} & \textbf{WG} & \textbf{OBQA} & \textbf{MMLU} & \textbf{GSM8K} & \textbf{HE+} & \textbf{MBPP+} & \textbf{Average} \\
\midrule
Instella-MoE-16B-A3B-Pretrain & 93.96 & 83.47 & 86.30 & 93.21 & 79.45 & 82.38 & 85.91 & 80.00 & 68.71 & 70.92 & 55.95 & 53.15 & 77.78 \\
Instella-MoE-16B-A3B-Midtrain-v1 & 93.84 & 84.90 & 84.90 & 93.71 & 82.00 & 82.09 & 87.09 & 83.60 & 68.95 & 77.82 & 56.86 & 55.00 & 79.23 \\
Instella-MoE-16B-A3B-Midtrain-v2 & 94.88 & 85.60 & 84.80 & 94.22 & 83.93 & 81.50 & 86.41 & 83.80 & 69.31 & 71.21 & 59.30 & 54.30 & 79.11 \\
Instella-MoE-16B-A3B-Midtrain-v3 & 94.63 & 85.83 & 86.00 & 94.45 & 85.19 & 81.26 & 85.85 & 84.40 & 69.69 & 75.93 & 58.54 & 54.95 & 79.73 \\
\rowcolor{InstellaRowBlue}
Instella-MoE-16B-A3B-Midtrain & 95.11 & 87.26 & 86.50 & 94.55 & 86.03 & 82.27 & 87.42 & 86.20 & 70.90 & 77.05 & 59.91 & 57.60 & 80.90 \\
\bottomrule
\end{tabular}%
}
\label{tab:midtraining_model_souping}
\end{table*}

\subsection{Long-Context Extension}
\label{sec:longcontext}
After mid-training, the context length of Instella-MoE-Midtrain is extended from 4K to 64K tokens through continued pre-training on AMD Instinct MI325X accelerators. The \emph{two-stage} recipe first adapts the model to the full 64K context length using a diverse set of long-context data; Stage~2 then rebalances the training mixture to recover mathematics, code, and reasoning capabilities that can regress during Stage~1. The checkpoint produced at the end of Stage~2 is \textbf{Instella-MoE-16B-A3B-Base} (Tables~\ref{tab:base_models} and~\ref{tab:longcontext}), and subsequent post-training stages initialize from this model and inherit the extended context window.

For long-context extension, we increase the RoPE base to $\theta=8\times10^{6}$ and apply YaRN scaling~\citep{peng2024yarn} to maintain attention stability at 64K. Starting with this stage, we also disable the MTP head to reduce activation memory at long sequence lengths. Ablations show that retaining $\theta=10{,}000$ without YaRN yields substantially worse length utilization; consequently, YaRN is used throughout both extension stages. Training samples are constructed by packing multiple documents into fixed 64K sequences. To prevent spurious cross-document dependencies, document masking is applied: attention is blocked across document boundaries and position indices restart at each new document.

\paragraph{Stage~1: Long-document adaptation.}
Stage~1 resumes from the soup-merged mid-training checkpoint and trains on \textbf{Dolma~3 Longmino 100B}~\citep{olmo2025olmo}, a mixture of OCR-derived science PDFs, long web pages, synthetic long-context aggregation tasks, and mid-training-quality subsets intended to limit catastrophic forgetting of general skills. Training uses ${\sim}$194B tokens at a sequence length of 64K, with a peak learning rate of $2\times10^{-4}$ and linear decay to zero. As shown in Table~\ref{tab:longcontext}, the resulting checkpoint, Instella-MoE-16B-A3B-Long-Stage1, uses long inputs effectively: across the 8K--64K sweep, it scores 44.7, 44.9, 44.4, and 41.0 on HELMET and averages 83.9 on RULER, outperforming the OLMo-3-7B model. Short-context STEM ability, however, regresses: GSM8K falls from 77.1 after mid-training to 62.8, and HumanEval+ falls from 59.9 to 55.0 (Tables~\ref{tab:base_models} and~\ref{tab:midtraining_model_souping}). Recovering this ability is the purpose of Stage~2.

\paragraph{Stage~2: STEM recovery.}
Stage~2 initializes from the Stage~1 checkpoint and runs for an additional ${\sim}$20B tokens with a peak learning rate of $1\times10^{-4}$ that decays linearly. Those tokens are sampled from a curated \textbf{37.32B-token STEM mixture} (Table~\ref{tab:longcontext_data}) built from Dolma~3 Dolmino~\citep{olmo2025olmo} and Instella-GSM8K-synthetic~\citep{liu2025instella}. The mixture includes instructional math, competition-style math, synthetic math reasoning, code corpora, code-reasoning traces, and short chain-of-thought rewrites. We pack samples into 64K sequences with the same document-boundary masking as Stage~1.

We evaluate four Stage~2 mixtures that vary the degree of STEM upsampling relative to general text. \textbf{General recovery} retains a broad distribution and improves GSM8K only modestly (64.5), with limited gains on coding. \textbf{Math recovery} increases math emphasis and restores GSM8K to 80.5, but incompletely recovers code performance (HumanEval+ 57.6). \textbf{STEM-intensive} further upweights math and code (GSM8K 81.0, HumanEval+ 64.5) with a trade-off in long-context length-sweep performance. \textbf{STEM-balanced} gives the best STEM recovery, restoring GSM8K to 81.5 and HumanEval+ to 65.7, at the cost of a modest long-context regression relative to Stage~1 (a decline in the HELMET average from 43.7 to 41.5 and in the RULER average from 83.9 to 79.4; Table~\ref{tab:longcontext}). We therefore adopt the STEM-balanced checkpoint as Instella-MoE-16B-A3B-Base.

\paragraph{Training Details.} Long-context extension at 64K on a 16B-parameter MoE is memory- and communication-intensive. The 64 routed experts are sharded across eight GPUs with expert parallelism, the sequence dimension is partitioned with context parallelism so each device holds a slice of the 64K activations, and full activation recomputation with bfloat16 mixed precision is used to fit a micro-batch size of one per step. Unlike the earlier 4K-context stages, the 64K-context training does not overlap forward and backward communication, and optimizer state is reinitialized for continued training at the new sequence length. Table~\ref{tab:longctx_hparams} summarizes the training hyperparameters.

\begin{table}[t]
    \centering
    \small
    \caption{\textbf{Long-context extension hyperparameters.} Both stages use a sequence length of 64K with document-boundary masking and YaRN RoPE ($\theta=8\times10^{6}$).}
    \label{tab:longctx_hparams}
    \setlength{\tabcolsep}{5pt}
    \begin{tabular}{@{}lcc@{}}
    \toprule
    & \textbf{Stage 1} & \textbf{Stage 2} \\
    \midrule
    Initialization & Mid-training (soup-merged) & Stage~1 checkpoint \\
    Training data & Dolma~3 Longmino 100B & 37.32B STEM mixture \\
    Sequence length & 64K & 64K \\
    Global batch size & 128 & 128  \\
    Training steps & 23{,}071 & 2{,}400 \\
    Peak learning rate & $2\times10^{-4}$ & $1\times10^{-4}$ \\
    Learning rate schedule & Linear decay, 200-step warmup & Linear decay \\
    MTP head & disabled & disabled \\
    Expert parallelism & 8-way & 8-way \\
    \bottomrule
    \end{tabular}
\end{table}

\begin{table}[t]
    \centering
    \small
    \caption{\textbf{Stage~2 STEM mixture (37.32B tokens).} Subsets are drawn from Dolma~3 Dolmino 100B, the full Dolma~3 Dolmino pool, or Instella-GSM8K-synthetic.}
    \label{tab:longcontext_data}
    \setlength{\tabcolsep}{5pt}
    \begin{tabular}{@{}llr@{}}
    \toprule
    \textbf{Domain} & \textbf{Source} & \textbf{Tokens} \\
    \midrule
    \multirow{6}{*}{Mathematics}
    & Instructional math (Dolmino) & 10.7B \\
    & Competition math & 5.62B \\
    & Synthetic math (MegaMatt) & 3.88B \\
    & Short math reasoning (Mind) & 898M \\
    & Short math reasoning (PoT) & 241M \\
    & GSM8K-style synthetic data & 329M \\
    \midrule
    Code & Instructional and competition code & 10.0B \\
    \midrule
    \multirow{4}{*}{Reasoning}
    & General reasoning traces & 2.48B \\
    & Rewritten full-thought chains & 850M \\
    & Math meta-reasoning & 1.05B \\
    & Code meta-reasoning & 1.27B \\
    \midrule
    \multicolumn{2}{@{}l}{\textbf{Total}} & \textbf{37.32B} \\
    \bottomrule
    \end{tabular}
\end{table}
\subsection{Supervised Fine-Tuning}
\label{sec:sft}
We initialize the SFT model from the long-context base model and train it in two phases. The phase-1 mixture combines the general Dolci-Think-SFT-7B dataset~\citep{olmo2025olmo} ($\sim$2.27M records) with three targeted skill slices: 300K math samples from Nemotron-Cascade-2~\citep{yang2026nemotroncascade2}, $\sim$161K Python competitive-programming samples from Nemotron-SFT-Competitive-Programming-v2~\citep{nemotronsftcp2}, and $\sim$197K science samples from Nemotron-Cascade-2 (Table~\ref{tab:sft_data}). We convert every source to a shared \texttt{\{"messages": [...]\}} format, drop system messages, and retain assistant reasoning traces. We also drop any conversation longer than $32{,}768$ tokens when tokenized with the DeepSeek-V3 chat template. Training uses sequence packing at a context length of 32K with a peak learning rate of $1\times10^{-4}$ and runs for two epochs.

\begin{figure}[t]
    \centering
    \resizebox{\linewidth}{!}{\definecolor{ink}{HTML}{1C1C1E}
\definecolor{inkgray}{HTML}{6E6E73}
\definecolor{dotgray}{HTML}{BCBCC2}
\definecolor{hitgreen}{HTML}{4C9A5E}
\definecolor{attnblue}{HTML}{C8DCF5}
\definecolor{salmonfill}{HTML}{F6D3CE}
\definecolor{routedfill}{HTML}{DCDFF0}
\definecolor{moegreen}{HTML}{CBE7CF}
\definecolor{grayfill}{HTML}{F3F3F4}
\definecolor{gatefill}{HTML}{F2F3C1}

\begin{tikzpicture}[x=1pt, y=1pt, outer sep=0pt,
    t9/.style={inner sep=0pt, text=ink, font=\small},
    t8/.style={inner sep=0pt, text=ink, font=\footnotesize},
    t8g/.style={inner sep=0pt, text=inkgray, font=\footnotesize},
    t7g/.style={inner sep=0pt, text=inkgray, font=\scriptsize},
    arr/.style={-{Triangle[length=4.6pt,width=4pt]}},
    wire/.style={draw=ink, line width=1.1pt, line join=round,
                 rounded corners=2.5pt},
    flow/.style={wire, arr},
    tap/.style={draw=inkgray, line width=1pt, line join=round,
                rounded corners=2.5pt, dash pattern=on 3.5pt off 2.5pt},
    taparr/.style={tap, arr},
    obox/.style={draw=ink, fill=#1, line width=1.1pt, rounded corners=2.5pt,
                 inner xsep=6pt, inner ysep=4pt, align=center, t9},
    card/.style={draw=ink, fill=white, line width=1.1pt, rounded corners=3pt},
    bandframe/.style={draw=inkgray, line width=1pt, rounded corners=5pt,
                      dash pattern=on 4.5pt off 3.5pt},
    tab/.style={t9, fill=white, inner xsep=4pt, inner ysep=1pt, anchor=west},
    qmark/.style={draw=ink, fill=gatefill, line width=0.8pt, rotate=45,
                  minimum size=7pt, inner sep=0pt},
    wbar/.style={draw=ink, fill=gatefill, line width=0.7pt, anchor=west,
                 minimum height=7pt, inner sep=0pt},
]

\draw[bandframe] (0,230) rectangle (500,322);
\node[tab] at (14,322) {\bfseries (i) Diagnosis};

\node[obox=grayfill, minimum width=84pt, minimum height=32pt] (seed) at (50,286)
     {Seed set\\{\footnotesize held-out prompts}};
\node[obox=attnblue, minimum width=84pt, minimum height=32pt] (stud) at (152,286)
     {Student\\{\footnotesize SFT checkpoint}};
\node[obox=salmonfill, minimum width=84pt, minimum height=32pt] (judge) at (276,286)
     {Judge\\{\footnotesize LLM-as-judge}};

\draw[flow] (seed.east) -- (stud.west);
\draw[flow] (stud.east) -- (judge.west);
\node[t8] at (214,295) {answers};

\draw[tap] (seed.south) -- (50,252) -- (276,252);
\draw[taparr] (276,252) -- (judge.south);
\node[t8g] at (163,244) {references};

\draw[tap] (judge.north) -- (276,314) -- (228,314);
\node[t8g, anchor=east] at (224,314) {correct $\to$ discarded};

\node[card, minimum width=154pt, minimum height=62pt] (ea) at (415,275) {};
\node[t8, anchor=west] at (346,298) {\bfseries Structured error analysis};
\node[t8, anchor=west] at (346,284) {\emph{error}: failure mode};
\node[t8, anchor=west] at (346,268) {\emph{skills}: missing capabilities};
\node[t8, anchor=west] at (346,252) {\emph{would help}: training data to retrieve};
\draw[flow] (judge.east) -- (338,286);
\node[t8] at (328,313) {incorrect};

\draw[bandframe] (0,134) rectangle (500,222);
\node[tab] at (14,222) {\bfseries (ii) Query generation};

\node[obox=routedfill, minimum width=118pt, minimum height=32pt] (refl) at (415,178)
     {Reflection model\\{\footnotesize aggregates error analyses}};
\fill[white] (409,217) rectangle (421,235);
\draw[flow] (415,244) -- (refl.north);

\node[card, minimum width=298pt, minimum height=76pt] (pol) at (173,178) {};
\node[t8, anchor=west] at (32,208) {\bfseries Retrieval policy};
\node[t8, anchor=east] at (314,208) {$\sum_i w_i = 1$};

\node[wbar, minimum width=40pt] at (34,191) {};
\node[wbar, minimum width=36pt] at (34,176) {};
\node[wbar, minimum width=10pt] at (34,161) {};
\node[t8, anchor=west] at (80,191) {0.020};
\node[t8, anchor=west] at (80,176) {0.018};
\node[t8, anchor=west] at (80,161) {0.005};
\node[t8, anchor=west] at (106,191) {Graph algorithms: BFS/DFS/Dijkstra on state spaces};
\node[t8, anchor=west] at (106,176) {Strict input/output format compliance};
\node[t8, anchor=west] at (106,161) {Spatial connectivity on geometric grids};
\node[t8] at (44,150) {$\vdots$};
\node[t7g, anchor=west] at (106,150) {one policy per targeted domain (code-domain policy shown)};

\draw[flow] (refl.west) -- (326,178);

\draw[bandframe] (0,0) rectangle (500,126);
\node[tab] at (14,126) {\bfseries (iii) Retrieval};

\fill[white] (134,124) rectangle (146,136);
\draw[flow] (140,140) -- (140,116);

\node[draw=ink, fill=grayfill, line width=1.1pt, rounded corners=3pt,
      minimum width=150pt, minimum height=82pt] at (99,75) {};
\node[t8] at (99,108) {Embedded candidate pool};

\foreach \p in {(32,98),(34,64),(36,44),(70,98),(96,96),(146,98),(160,86),(152,68),
                (164,50),(140,44),(118,56),(104,66),(100,84),(76,66),(64,60),(44,54),
                (38,80),(130,98),(156,40),(134,64),(108,44),(70,40),(50,40),(166,98),
                (28,50),(88,78),(146,80)}
  \fill[dotgray] \p circle[radius=1.7pt];

\foreach \c/\r in {(56,86)/14, (122,80)/14, (84,50)/12}
  \draw[draw=inkgray, line width=0.8pt, dash pattern=on 2pt off 1.6pt]
        \c circle[radius=\r pt];
\foreach \p in {(48,90),(62,92),(52,78),(64,80),(114,86),(128,86),(116,74),(130,76),
                (78,54),(90,54),(84,42)}
  \fill[hitgreen] \p circle[radius=1.9pt];
\foreach \c in {(56,86),(122,80),(84,50)} \node[qmark] at \c {};

\node[qmark] at (30,20) {};
\node[t8, anchor=west] at (37,20) {query $q_i$};
\fill[hitgreen] (98,20) circle[radius=1.9pt];
\node[t8, anchor=west] at (104,20) {retrieved neighbor};

\draw[wire] (174,80) -- (290,80);
\draw[wire] (290,80) -- (290,100);
\draw[flow] (290,100) -- (304,100);
\draw[flow] (290,80) -- (304,80);
\fill[ink] (290,80) circle[radius=1.8pt];
\node[t8] at (238,96) {top-$k_i$ nearest neighbors};
\node[t8] at (238,87) {(cosine similarity)};
\node[t8] at (238,69) {$k_i=\lfloor B_{\mathrm{dom}}\,w_i\,(1{-}\alpha)\rfloor$};

\foreach \y/\xmid/\xend in {100/379/452, 80/364.5/423}{
  \draw[draw=ink, fill=moegreen, line width=1pt] (306,\y-7) rectangle (\xmid,\y+7);
  \draw[draw=ink, fill=grayfill, line width=1pt] (\xmid,\y-7) rectangle (\xend,\y+7);
}
\draw[draw=ink, fill=grayfill, line width=1pt] (306,53) rectangle (414.5,67);
\node[t8, anchor=west] at (458,100) {Math};
\node[t8, anchor=west] at (458,80) {Code};
\node[t8, anchor=west] at (458,60) {Other};
\node[t8] at (342,116) {retrieved ($1{-}\alpha$)};
\node[t8] at (415,116) {uniform fill ($\alpha$)};
\node[t7g, anchor=east] at (452,44) {gray: uniform same-domain fill, $\alpha{=}0.5$};

\node[obox=moegreen, minimum width=214pt, minimum height=24pt] (out) at (379,24)
     {Curated 512K mixture $\to$ final SFT phase};
\draw[flow] (316,53) -- (316,36);

\end{tikzpicture}}
    \caption{\textbf{Feedback-driven data curation for the final SFT phase.} \emph{(i)} The student answers questions from a held-out seed set, and a judge model scores each response against the reference answer; correct responses are dropped, and each failure is turned into a structured error analysis naming the failure mode, the missing skills, and the training data that would close the gap. \emph{(ii)} A larger reflection model aggregates the analyses for a domain into a retrieval policy: a set of queries with weights summing to one. \emph{(iii)} Each query retrieves its allocated $k_i$ examples from the per-domain budget $B_{\mathrm{dom}}$ through nearest-neighbor search in the shared embedding space of the candidate pool. Uniform same-domain sampling fills the remaining $\alpha$ share of each targeted-domain budget and the entire budget for the non-targeted \emph{other} domain. The total budget $B=512$K is split across domains in the proportions of the base mixture, as reflected by the differing bar lengths.}
    \label{fig:sft_curation}
\end{figure}
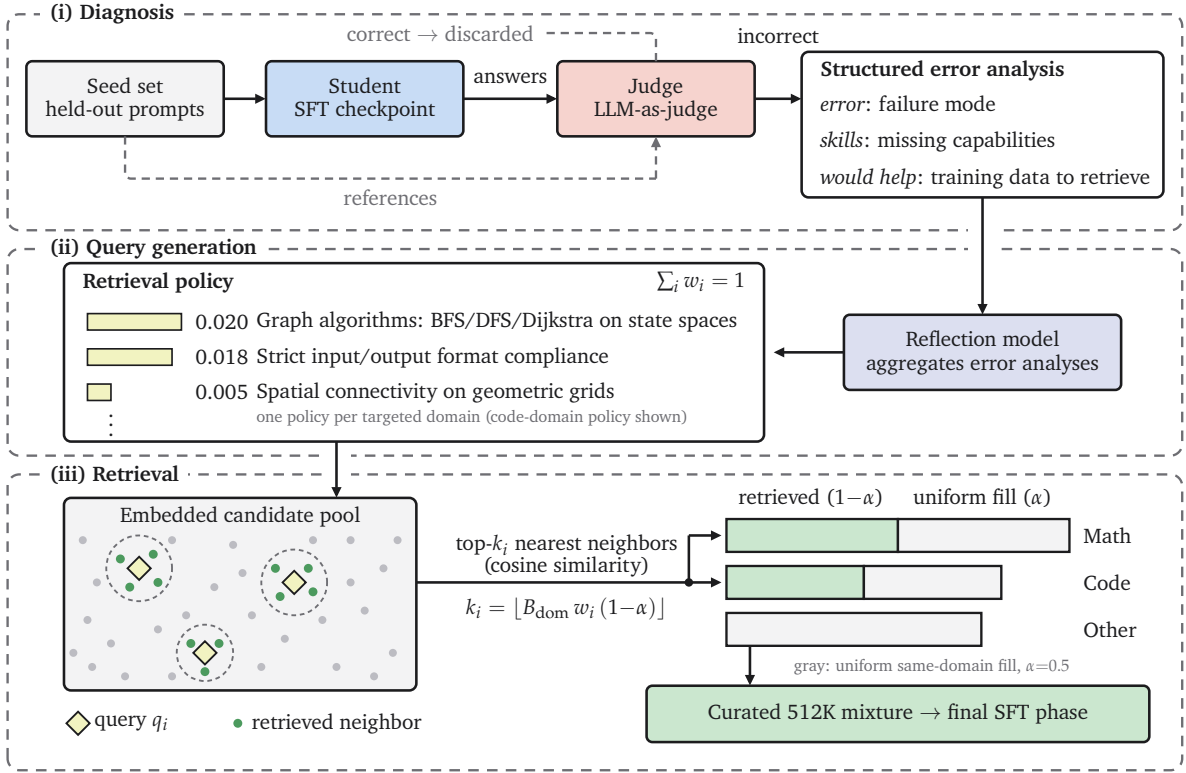

\begin{figure}[t]
    \centering
    \includegraphics[width=0.9\linewidth]{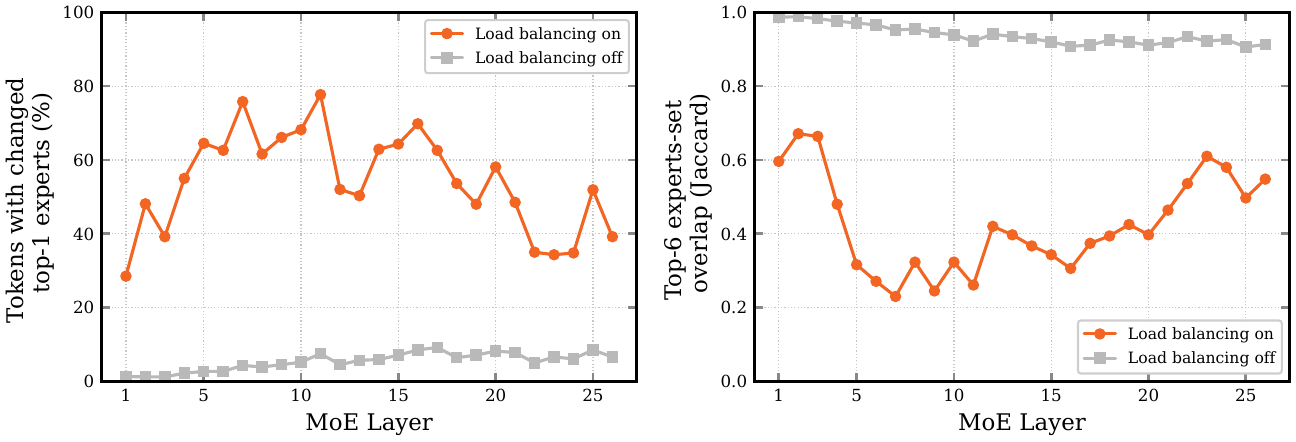}
    \caption{\textbf{Expert routing drift during DPO}. Two otherwise identical DPO runs, differing only in whether load balancing remains enabled, are compared against the SFT model across 26 MoE layers. \emph{Left:} fraction of tokens whose top-1 expert differs from SFT. \emph{Right:} Jaccard overlap between top-6 dispatched experts and those of SFT. Routing stays close to SFT when load balancing is disabled, and differs substantially when enabled.}
    \label{fig:routing_drift}
\end{figure}

\textbf{Feedback-driven data curation.} Once gains from the core
mixture saturate, we run a final phase on 512K examples selected by an
error-driven procedure rather than by uniform sampling. We partition
the pool into three domains---mathematics, code, and other---and split a fixed
selection budget across them. For the two targeted domains,
mathematics and code, we select data using a three-stage pipeline
(Figure~\ref{fig:sft_curation}). \emph{(i) Diagnosis:} we run the
student on a held-out seed set and use an LLM as a judge to score each response
against a reference answer; for every incorrect response, the judge model then
produces a structured error analysis that includes a description of what
training data would help address the error. \emph{(ii) Query generation:} a larger reflection
model aggregates these error analyses into a set of weighted retrieval queries,
with weights normalized to sum to one. \emph{(iii) Retrieval:} we allocate each query its weighted share of
the selection budget and retrieve that many nearest neighbors from the pool by embedding similarity. Compared to uniform
sampling at the same budget, this improves average performance across our
evaluation suite by $1.5$ points (Table~\ref{tab:sft_curation_ablation}),
with the largest gains on instruction following, competition math, and
code generation. A parallel line of work,
STAT~\citep{stat}, also targets a student's weaknesses through teacher-diagnosed
errors. It relies on a predefined skill taxonomy to drive selection. In our
method, the teacher provides free-form error feedback, which we convert into
retrieval queries over a large candidate pool. More details about the
implementation are provided in Appendix~\ref{app:sft_curation}.

\begin{table}[t]
\centering
\setlength{\tabcolsep}{3pt}
\renewcommand{\arraystretch}{1.1}
\scriptsize
\caption{\textbf{Final-phase SFT data selection.} Both runs anneal the same phase-1 model on 512K examples drawn from the same candidate pool and differ only in how those examples are selected. Feedback-driven curation improves the eleven-benchmark average by $1.5$ points, with the largest gains on IFEval ($+4.8$), AIME25 ($+3.6$), and LCB ($+3.2$). HE+ and LCB denote HumanEval+ and LiveCodeBench. Per column, the best score is bold.}
\resizebox{\columnwidth}{!}{%
\begin{tabular}{l|ccccccccccc|c}
\toprule
\textbf{Selection} & \textbf{AGIEval} & \textbf{AIME24} & \textbf{AIME25} & \textbf{BBH} & \textbf{HE+} & \textbf{GPQA} & \textbf{IFEval} & \textbf{LCB} & \textbf{MBPP+} & \textbf{MATH} & \textbf{MMLU} & \textbf{Avg} \\
\midrule
Uniform sampling & 79.86 & \textbf{78.02} & 67.71 & 74.04 & 85.45 & 58.00 & 76.71 & 45.53 & 61.59 & \textbf{94.77} & 79.70 & 72.85 \\
\rowcolor{InstellaRowBlue}
Feedback-driven curation & \textbf{81.33} & 77.50 & \textbf{71.35} & \textbf{74.34} & \textbf{86.77} & \textbf{59.38} & \textbf{81.51} & \textbf{48.74} & \textbf{61.64} & 94.67 & \textbf{80.82} & \textbf{74.37} \\
\bottomrule
\end{tabular}%
}
\label{tab:sft_curation_ablation}
\end{table}

\subsection{Direct Preference Optimization}
\label{sec:dpo}

Starting from the curated SFT checkpoint, we apply DPO~\citep{dpo} using the Dolci-Think-DPO-7B preference dataset~\citep{olmo2025olmo}, which is built on the idea of \emph{delta learning}~\citep{deltalearning}: what drives preference tuning is the quality \emph{gap} between the chosen and rejected responses rather than the absolute quality of either, so each pair contrasts a Qwen3-32B thinking trace~\citep{yang2025qwen3} with a much weaker Qwen3-0.6B trace and remains informative even when continued imitation of the chosen traces alone would no longer provide a useful training signal. When run naively, DPO converges, yet the resulting model's accuracy is lower than that of the SFT checkpoint on downstream benchmarks. We do not observe this behavior when applying the same recipe to a dense model. We also note that the DPO stage operates at a substantially lower learning rate than pre-training and SFT, while the MoE auxiliary load-balancing loss and router bias updates are configured for those higher-learning-rate regimes. Motivated by these observations, we disable both load-balancing mechanisms during DPO. As shown in Figure~\ref{fig:routing_drift}, when load balancing is enabled, the top-1 expert differs from that of SFT for 54\% of tokens on average and the top-6 expert-set overlap is 0.42, compared with 5\% and 0.94, respectively, when load balancing is disabled. With these adjustments, DPO improves helpfulness and response quality while preserving reasoning.

\subsection{Reinforcement Learning}
\label{sec:rl}

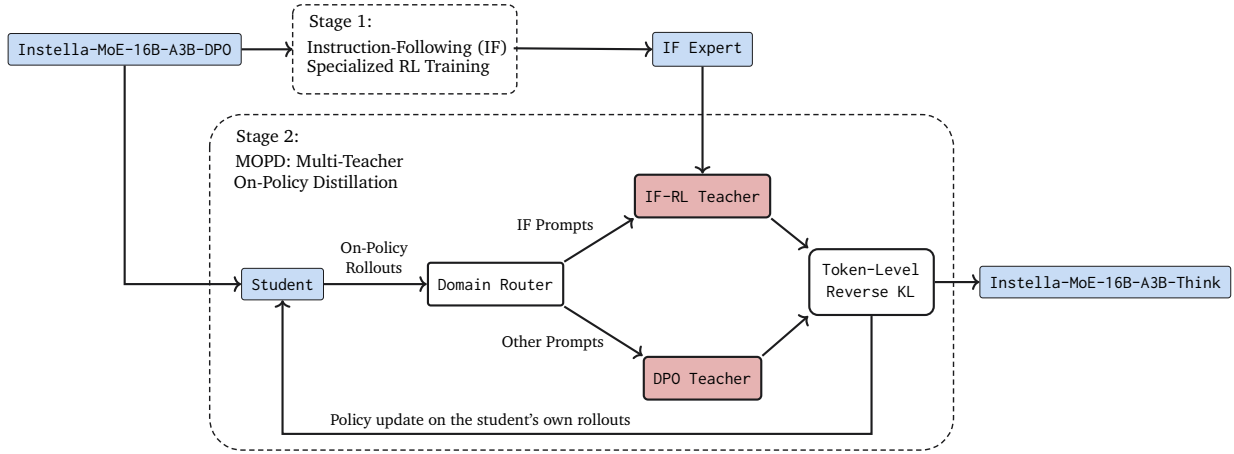
\begin{figure}[t]
    \centering
    \resizebox{\linewidth}{!}{\definecolor{ink}{HTML}{1C1C1E}
\definecolor{attnblue}{HTML}{C8DCF5}
\definecolor{mlared}{HTML}{E6B3B2}
\definecolor{moegreen}{HTML}{CBE7CF}

\begin{tikzpicture}

  \node[draw=ink, text=black, fill=attnblue, rounded corners=1.6pt, inner sep=5.53pt, node font=\ttfamily] (node1) at (-0.53,3) {Instella-MoE-16B-A3B-DPO};
  \draw[rounded corners=5.3pt, ink, thick, fill=white, densely dashed] (2.79,3.88) rectangle (7.22,2.12);
  \draw[rounded corners=18.1pt, ink, thick, fill=white, densely dashed] (1.15,1.71) rectangle (15.85,-4.92);
  \node[draw=none, inner sep=3.13pt] (node5) at (3.32,0.8) {{MOPD: Multi-Teacher}};
  \node[draw=none, inner sep=3.13pt] (node6) at (3.25,0.36) {{On-Policy Distillation}};
  \draw[->, very thick, ink] (node1.east) -- (2.79,3);
  \node[draw=none, align=left, node font=] at (3.74,3.53) {Stage 1:};
  \node[draw=none, align=left, node font=] at (2.3,1.24) {Stage 2:};
  \node[draw=none, inner sep=3.13pt] (node3) at (5.06,3.02) {{Instruction-Following (IF)}};
  \node[draw=none, inner sep=3.13pt] (node4) at (4.88,2.64) {{Specialized RL Training}};

  \node[draw=ink, text=black, fill=attnblue, rounded corners=1.6pt, inner sep=5.53pt, node font=\ttfamily] (node2) at (10.89,3) {IF Expert};
  \node[draw=ink, text=black, fill=attnblue, rounded corners=1.6pt, inner sep=5.53pt, node font=\ttfamily] (node7) at (2.59,-1.64) {Student};

  \draw[->, very thick, ink] (node3.east) -- (node2.west);

  \draw[->,very thick, ink] (node1.south) -- (-0.53,-1.64) -- (node7.west);

  \node[draw=ink, text=black, fill=white, rounded corners=1.6pt, inner sep=5.53pt, node font=\ttfamily, minimum height=24pt, very thick] (node8) at (6.8,-1.64) {Domain Router};
  \node[draw=ink, text=black, fill=white, rounded corners=4.9pt, inner sep=5.53pt, node font=\ttfamily, minimum height=37pt, very thick, minimum width=70pt] (node15) at (14.23,-1.6) {};
  \node[draw=ink, fill=mlared, text=black, rounded corners=2.4pt, inner sep=5.53pt, node font=\ttfamily, minimum height=24pt, very thick] (node11) at (10.89,0.1) {IF-RL Teacher};
  \node[draw=ink, fill=mlared, text=black, rounded corners=2.4pt, inner sep=5.53pt, node font=\ttfamily, minimum height=24pt, very thick] (node12) at (10.89,-3.5) {DPO Teacher};
  \draw[->, very thick, ink] (node7.east) -- (node8.west);

  \node[draw=none, inner sep=3.13pt, node font=\footnotesize] (node9) at (4.4,-0.95) {{On-Policy}};
  \node[draw=none, inner sep=3.13pt, node font=\footnotesize] (node13) at (7.98,-0.49) {{IF Prompts}};
  \node[draw=none, inner sep=3.13pt, node font=\footnotesize] (node14) at (7.93,-2.8) {{Other Prompts}};

  \node[draw=none, inner sep=3.13pt, node font=\footnotesize] (node10) at (4.4,-1.3) {{Rollouts}};
  \draw[->,very thick, ink] (node8.north east) -- (node11.south west);
  \draw[->,very thick, ink] (node8.south east) -- (node12.north west);
  \draw[->,very thick, ink] (node2.south) -- (node11.north);

  \node[draw=none, inner sep=3.13pt, node font=\ttfamily] (node16) at (14.22,-1.35) {{Token-Level}};
  \node[draw=none, inner sep=3.13pt, node font=\ttfamily] (node17) at (14.22,-1.78) {{Reverse KL}};
  \draw[->,very thick, ink] (node11.south east) -- (node15.north west);
  \draw[->,very thick, ink] (node12.north east) -- (node15.south west);
  \draw[->, very thick, ink] (node15.south) -- (14.23,-4.6) -- (2.59,-4.6) -- (node7.south);

  \node[draw=none, inner sep=3.13pt, node font=\footnotesize] (node18) at (6.5,-4.35) {{Policy update on the student's own rollouts}};

  \node[draw=ink, text=black, fill=attnblue, rounded corners=1.6pt, inner sep=5.53pt, node font=\ttfamily] (node19) at (18.85,-1.6) {Instella-MoE-16B-A3B-Think};
  \draw[->,very thick, ink] (node15.east) -- (node19.west);
\end{tikzpicture}}
    \caption{\textbf{Instella-MoE-16B-A3B RL training pipeline.} Stage~1 performs IF-specialized RL to obtain an IF expert. Stage~2 applies Multi-Teacher On-Policy Distillation (MOPD): a domain router routes the student's IF rollouts to the IF-RL teacher and all other rollouts to the frozen DPO teacher, and the student is updated on its own on-policy rollouts using a token-level reverse-KL objective.}
    \label{fig:rl_pipeline}
\end{figure}

To obtain our final Think checkpoint, we apply reinforcement learning (RL) to the DPO
model using Miles~\citep{radixark2025miles, slime_github}. Because the DPO model already performs strongly
on math and general reasoning, we restrict the RL objective to
\emph{instruction following} (IF) and then use an on-policy distillation stage to incorporate those gains
into the DPO model without eroding its other capabilities. The two-stage recipe
is illustrated in Figure~\ref{fig:rl_pipeline}: Stage~1 is IF-specialized RL with
verifiable rewards (RLVR), which produces an IF expert, and Stage~2 is
Multi-Teacher On-Policy Distillation (MOPD), which distills knowledge from that
expert and the DPO model into the student, with the DPO model acting as a self-anchor.

\subsubsection{RL Infrastructure}
Both stages use a disaggregated actor--learner design built on Miles~\citep{radixark2025miles, slime_github}:
rollouts are produced by a
pool of SGLang~\citep{zheng2024sglang} inference engines, while the policy is optimized in Megatron-LM~\citep{shoeybi2019megatron}.
After every optimizer step, the updated weights are streamed back to the rollout
engines. We run the IF-RL stage
(Section~\ref{par:if-rl}) \emph{fully asynchronously}, which allows generation to overlap with
gradient computation in a single-step off-policy regime. For the MOPD stage
(Section~\ref{par:mopd}), we train \emph{synchronously and on-policy}, since the teacher
scoring makes each update depend on the current student. Each run uses three nodes
with 24 GPUs, split into 8 GPUs for the learner (expert parallelism, EP${=}8$, with a distributed optimizer)
and 16 GPUs for two parallel SGLang rollout engines (each TP${=}8$, EP${=}8$, with
data-parallel attention). The rollout engines generate responses of up to 16K tokens.

\paragraph{MoE-aware train/inference consistency.}
A subtlety specific to the MoE architecture is that the SGLang rollout and the Megatron
training forward pass can route tokens to different experts, so the training-time
log-probabilities $\log\pi^{\text{train}}$ need not match those under which the
tokens were sampled. We address this at two levels. First, we enable Rollout
Routing Replay (R3)~\citep{ma2025stabilizingmoereinforcementlearning}, which fixes the expert-assignment decisions made during
sampling by recording and replaying them inside the training forward pass. With this
approach, $\log\pi^{\text{train}}$ is evaluated along the exact expert path that generated
each token. Second, we apply token-level truncated importance sampling (TIS)~\citep{yao2025offpolicy} to correct residual numerical mismatch between the inference and training engines, with per-token ratios clamped to a bounded range (Table~\ref{tab:rl-hparams}).

\subsubsection{IF-specialized RL}
\label{par:if-rl}
In the first stage, we train the DPO model for $1{,}400$ rollout steps on the IF-RLVR subset of
Dolci-Think-RL-7B~\citep{olmo2025olmo}, whose prompts carry verifiable instruction-following constraints
in the style of IFEval~\citep{zhou2023instructionfollowingevaluationlargelanguage} and IFBench~\citep{pyatkin2025generalizing}. Each step samples
$64$ prompts $\times\,8$ responses $=512$ rollouts (global batch size $512$),
decoding at temperature $1.0$ with a maximum response length of $16{,}384$ tokens.
We additionally use \emph{partial rollouts}~\citep{zhou2025aprilactivepartialrollouts}: long-tail decodes that do not finish
within a step are recycled into the next step, with the stale (off-policy) tokens
masked out of the loss. This approach substantially improves throughput at these response
lengths. For this stage, we use Adam ($\beta_1{=}0.9$, $\beta_2{=}0.999$) with a constant
learning rate of $1\times10^{-6}$, weight decay $0$, and gradient clipping of $1.0$,
and monitor IFEval and IFBench every $50$ steps (evaluated with a temperature of $0.6$
and top-$p$ of $0.95$). This yields an \emph{IF expert} with markedly stronger
instruction-following behavior than our DPO model.

Our core objective is Group Relative Policy Optimization (GRPO)~\citep{shao2024deepseekmathpushinglimitsmathematical},
augmented with a set of now-standard improvements drawn from DAPO~\citep{yu2025dapoopensourcellmreinforcement} and
Dr.~GRPO~\citep{liu2025understanding}. For a prompt $x$, we sample a group of $G$ responses
$\{y_i\}_{i=1}^{G}$ from the rollout policy and optimize
\[
\mathcal{J}(\theta)=
\frac{1}{\sum_{i=1}^{G}|y_i|}
\sum_{i=1}^{G}\sum_{t=1}^{|y_i|}
w_{i,t}\,
\min\!\Big(
r_{i,t}\,A_{i,t},\;
\mathrm{clip}\big(r_{i,t},\,1-\varepsilon_{\text{low}},\,1+\varepsilon_{\text{high}}\big)\,A_{i,t}
\Big),
\]
where $r_{i,t}=\dfrac{\pi_\theta(y_{i,t}\mid x,y_{i,<t})}{\pi_{\theta_{\text{old}}}(y_{i,t}\mid x,y_{i,<t})}$
is the policy ratio, $w_{i,t}$ is the truncated importance-sampling weight that
corrects the mismatch between the rollout and learner policies, and the advantage is computed
\emph{group-relative} with mean-centering only,
$A_{i,t}= R(x,y_i)-\operatorname{mean}(\{R(x,y_j)\}_{j=1}^{G})$, where $R$ is the
verifier reward described below.
The specific modifications to vanilla GRPO, all of which we use, are:
\begin{itemize}
  \item \textbf{Zero-gradient signal filtering.} Groups in which all $G$
  responses receive identical reward have zero advantage and contribute no
  gradient; we discard them~\citep{yu2025dapoopensourcellmreinforcement}.
  \item \textbf{Active sampling.} To keep a constant, fully informative batch
  size despite this filtering, we oversample prompts (128 per step) and continue
  sampling until the batch is filled with groups with nonzero reward variance,
  exploiting the asynchronous rollout pool so the learner never waits.
  \item \textbf{Token-level loss.} The loss is normalized by the total number of
  tokens in the batch rather than being normalized per sequence, removing the length bias that
  underweights long reasoning traces~\citep{yu2025dapoopensourcellmreinforcement}.
  \item \textbf{No KL penalty.} We drop the KL-to-reference term (and reference
  model) entirely, which permits larger policy movement without destabilizing
  training~\citep{yu2025dapoopensourcellmreinforcement,liu2025understanding}.
  \item \textbf{Clip-higher.} We decouple the PPO clip bounds, using
  $\varepsilon_{\text{low}}=0.2$ and $\varepsilon_{\text{high}}=0.272$ to preserve
  probability mass on exploratory low-probability tokens~\citep{yu2025dapoopensourcellmreinforcement}.
  \item \textbf{No standard-deviation normalization.} Advantages are
  mean-centered but not divided by the group reward standard deviation, removing the difficulty
  bias whereby very easy or very hard prompts receive inflated
  advantages~\citep{liu2025understanding}.
  \item We use \textbf{Truncated importance sampling (TIS)} and \textbf{Rollout Routing
  Replay (R3)}~\citep{yao2025offpolicy,ma2025stabilizingmoereinforcementlearning} for train/inference consistency, as described
  above.
\end{itemize}

\paragraph{Verifier-based reward design.} Rewards are scaled to the $[0,10]$ range and are produced by deterministic verifiers rather than a learned reward model. This design avoids reward hacking on long reasoning traces. Instruction-following responses are graded by a sequence of constraint-checking functions. Following the IF-RLVR recipe of \citet{pyatkin2025generalizing} (also adopted by OLMo~3~\citep{olmo2025olmo}), we assign \emph{partial} credit as the fraction of constraints satisfied rather than as an all-or-nothing score. This enriches the training signal and eases credit assignment on multi-constraint prompts. The IF constraint verifier drives Stage~1 training and is also used for evaluation (IFEval/IFBench); during evaluation, however, we adopt the \emph{strict} criterion, awarding credit only when \emph{all} of a prompt's constraints are satisfied. MOPD replaces the scalar reward with a teacher-log-probability signal and therefore does not use these verifiers during training. Before any verifiable credit is assigned, we apply a \emph{think-format gate}: a response earns a reward only if it contains a well-formed reasoning block ($\texttt{<think>}\dots\texttt{</think>}$); the reasoning trace is then stripped and only the final-answer span is scored, preventing malformed or non-reasoning outputs from receiving credit.

\subsubsection{Multi-Teacher On-Policy Distillation}
\label{par:mopd}
IF-specialized RL yields an IF expert that excels at instruction following but
regresses on math and reasoning benchmarks (Table~\ref{tab:rl_ablation}). We apply MOPD~\citep{ma2026mopd}
to integrate the instruction-following capability of the IF expert into the DPO model while
preserving the DPO model's other capabilities. Specifically, we distill on the student's own
on-policy rollouts over a domain-tagged mixture of Dolci-Think-RL-7B prompts (an
even split between IF and general prompts; $128$ prompts $\times\,4$ samples $=512$ responses per step;
maximum response length 16K tokens), routing each rollout to a \emph{frozen} teacher. Each teacher is served through an
independent SGLang endpoint; the prompt's domain determines which teacher scores the
student's sampled tokens. Instruction-following prompts are scored by the IF-RL teacher,
while all remaining prompts (e.g., math, code, and general) are scored by the frozen DPO
model itself. The student is initialized from the DPO model, so on the non-IF
domains the teacher is the student's own initialization, which results in an advantage
that is approximately zero early in training and acts as a self-anchoring regularizer
that preserves math and general capability while the IF-domain signal drives learning.
Concretely, we adopt an on-policy distillation objective in which the per-token
advantage is the teacher--student log-probability gap:
\begin{equation}
A_t = \log \pi_{\text{teacher}}(y_t \mid y_{<t}, x) - \log \pi_{\text{student}}(y_t \mid y_{<t}, x).
\end{equation}
The expectation of this gap under the student's on-policy distribution corresponds to a token-level reverse-KL objective between the student and the domain-routed teacher.
Teacher log-probabilities are obtained by scoring the student's sampled tokens (no
teacher generation), and the domain router selects $\pi_{\text{teacher}}$ per
sample from $\{\pi_{\text{IF-RL}},\, \pi_{\text{DPO}}\}$. This objective is optimized within
the same GRPO/importance-sampling machinery (importance ratios truncated to
$[0.5,2.0]$, no KL term, and no entropy bonus). As shown in Table~\ref{tab:rl_ablation}, MOPD
recovers most of the IF expert's instruction-following gains while maintaining
DPO-level performance on math, code, MMLU~\citep{hendrycks2021measuringmassivemultitasklanguage}, and AGIEval~\citep{zhong2024agieval}.
\begin{table}[t]
\centering
\small
\caption{\textbf{Reinforcement learning hyperparameters for the two Think RL stages.}}
\label{tab:rl-hparams}
\begin{tabular}{lcc}
\toprule
& \textbf{Stage 1: IF-RL} & \textbf{Stage 2: MOPD} \\
\midrule
Objective            & GRPO (verifiable IF reward)        & On-policy distillation (reverse KL) \\
Initialization       & DPO checkpoint                     & DPO checkpoint \\
Data                 & Dolci-Think-RL IF-RLVR subset      & Dolci-Think-RL IF/general mix (50\% IF) \\
Prompts $\times$ samples & $64\times 8$                   & $128\times 4$ \\
Global batch size    & $512$                              & $512$ \\
Max response length  & $16{,}384$                         & $16{,}384$ \\
Sampling             & $T{=}1.0$ (top-$p$ default)        & $T{=}1.0$, top-$p\,{=}\,1.0$ \\
Learning rate        & $1\times10^{-6}$ (constant)        & $1\times10^{-6}$ (constant, 30-step warmup) \\
Adam $(\beta_1,\beta_2)$ & $(0.9,\,0.999)$                & $(0.9,\,0.95)$ \\
Clip $(\varepsilon_{\text{low}},\varepsilon_{\text{high}})$ & $(0.2,\,0.272)$ & --- \\
Importance clip (TIS) & $[0.5,\,1.5]$                     & $[0.5,\,2.0]$ \\
KL / entropy         & $0$ / $0$                          & $0$ / $0$ \\
MoE routing          & R3 replay                          & R3 replay \\
Rollout mode         & asynchronous, partial rollouts     & synchronous, on-policy \\
\bottomrule
\end{tabular}
\end{table}

\section{Results}

\subsection{Base Model}
\begin{table*}[t]
\centering
\setlength{\tabcolsep}{3pt}
\renewcommand{\arraystretch}{1.1}
\scriptsize
\caption{\textbf{Base model performance on standard pre-training benchmarks.} HSwag, WG, OBQA, HE+, and MATH denote HellaSwag, WinoGrande, OpenBookQA, HumanEval+, and Minerva MATH, respectively. \emph{Long-Stage1} is the intermediate checkpoint after long-context Stage~1, before STEM recovery. Per column, the best score is bold and the second-best is underlined.}
\resizebox{\textwidth}{!}{%
\begin{tabular}{l|*{14}{c}}
\toprule
\textbf{Model} & \textbf{ARC-E} & \textbf{ARC-C} & \textbf{BoolQ} & \textbf{SciQ} & \textbf{PIQA} & \textbf{HSwag} & \textbf{WG} & \textbf{OBQA} & \textbf{MMLU} & \textbf{GSM8K} & \textbf{HE+} & \textbf{MBPP+} & \textbf{MATH} & \textbf{Avg} \\
\midrule
\multicolumn{15}{c}{\textit{Open-Weight Models}} \\
\midrule
Llama-3.2-3B & 85.3 & 71.2 & 74.2 & 87.1 & 74.1 & 77.0 & 83.0 & 67.0 & 57.5 & 30.2 & 31.0 & 32.4 & 8.7 & 59.9 \\
Gemma-4-E2B & 53.2 & 41.7 & 78.9 & 39.6 & 53.4 & 47.5 & 61.2 & 74.8 & 59.2 & 24.0 & 22.0 & 31.0 & 9.4 & 45.8 \\
Gemma-4-E4B & 48.8 & 37.9 & \underline{87.2} & 43.5 & 50.2 & 52.9 & 63.4 & \underline{85.2} & 70.9 & 56.4 & 44.5 & 42.5 & 20.6 & 54.1 \\
Qwen3.5-2B-Base & 93.7 & 84.4 & 85.8 & 91.9 & 76.7 & 68.4 & 78.3 & 77.4 & 65.5 & 67.2 & 34.9 & 37.8 & 32.5 & 68.8 \\
Qwen3.5-4B-Base & \textbf{97.5} & \textbf{93.1} & \textbf{89.8} & \textbf{95.4} & \textbf{87.1} & 77.1 & 84.6 & \textbf{89.4} & \textbf{77.3} & \underline{84.4} & \underline{55.8} & \textbf{53.0} & \textbf{48.3} & \textbf{79.5} \\
DeepSeek-V2-Lite & 87.6 & 74.8 & 81.7 & 88.3 & 71.4 & \underline{80.8} & 84.8 & 72.4 & 58.8 & 38.9 & 32.5 & 32.4 & 15.6 & 63.1 \\
Moonlight-16B-A3B & \underline{95.3} & \underline{85.4} & 85.9 & \underline{94.2} & 81.8 & \textbf{82.5} & 85.6 & 82.2 & \underline{71.3} & \textbf{85.6} & 52.1 & 45.7 & \underline{43.0} & 76.2 \\
\midrule
\multicolumn{15}{c}{\textit{Fully Open Models}} \\
\midrule
OLMoE-1B-7B & 84.4 & 69.2 & 83.9 & 87.0 & 73.5 & 79.9 & 84.1 & 79.0 & 62.5 & 51.0 & 14.5 & 20.4 & 15.2 & 61.9 \\
SmolLM3-3B-Base & 90.9 & 79.8 & 83.8 & 90.1 & 73.9 & 77.8 & 84.4 & 77.4 & 62.6 & 69.7 & 37.8 & 49.2 & 39.6 & 70.5 \\
OLMo-3-7B & 93.6 & 84.8 & 63.9 & 92.8 & 80.2 & 77.7 & 85.7 & 69.6 & 56.4 & 74.6 & 48.0 & 43.8 & 40.2 & 70.1 \\
\rowcolor{InstellaRowBlue}
Instella-MoE-16B-A3B-Long-Stage1 & 93.6 & 83.9 & 84.1 & 93.3 & 82.7 & 80.1 & \underline{86.0} & 82.2 & 68.9 & 62.8 & 55.0 & \underline{52.7} & 39.1 & 74.2 \\
\rowcolor{InstellaRowBlue}
Instella-MoE-16B-A3B-Base & 93.6 & 83.8 & 84.7 & 93.6 & \underline{83.8} & 79.2 & \textbf{86.5} & 81.2 & 67.8 & 81.5 & \textbf{65.7} & 52.5 & \underline{43.0} & \underline{76.7} \\
\bottomrule
\end{tabular}%
}
\label{tab:base_models}
\end{table*}

We evaluate Instella-MoE-16B-A3B-Base on ARC-Challenge (ARC-C)~\citep{clark2018thinksolvedquestionanswering}, ARC-Easy (ARC-E)~\citep{clark2018thinksolvedquestionanswering}, BoolQ~\citep{clark-etal-2019-boolq}, HellaSwag (HSwag)~\citep{zellers-etal-2019-hellaswag}, PIQA~\citep{bisk2019piqareasoningphysicalcommonsense}, SciQ~\citep{welbl2017crowdsourcingmultiplechoicescience}, WinoGrande (WG)~\citep{sakaguchi2019winograndeadversarialwinogradschema}, OpenBookQA (OBQA)~\citep{mihaylov2018suitarmorconductelectricity}, MMLU~\citep{hendrycks2021measuringmassivemultitasklanguage}, GSM8K~\citep{cobbe2021trainingverifierssolvemath}, HumanEval+ (HE+) and MBPP+~\citep{evalplus}, and Minerva MATH~\citep{lewkowycz2022solving}. Unless otherwise noted, we follow the evaluation protocols described in our Hugging Face model cards and use OLMES-style settings for base models~\citep{gu-etal-2025-olmes}.

As shown in Table~\ref{tab:base_models} and Figure~\ref{fig:cost_performance} (left), Instella-MoE-16B-A3B-Base attains an average score of \textbf{76.7}, the strongest among all fully open models---well ahead of SmolLM3-3B~\citep{bakouch2025smollm3} (70.5), OLMo-3-7B~\citep{olmo2025olmo} (70.1), and OLMoE-1B-7B~\citep{olmoe} (61.9). It leads all evaluated models on WinoGrande (86.5) and delivers strong coding results (HumanEval+ 65.7), with balanced performance across knowledge, reasoning, math, and code. Notably, it achieves these results while activating only 2.8B parameters per token, outperforming fully open dense models such as OLMo-3-7B that activate more than twice as many parameters.

Instella-MoE is also highly competitive with leading open-weight models: it surpasses Moonlight-16B-A3B~\citep{liu2025muonscalablellmtraining} (76.2) on average while being fully open, and trails only Qwen3.5-4B-Base~\citep{qwen3.5} (79.5), which activates more parameters per token.

\paragraph{Long-Context Evaluation.}
\begin{table*}[t]
\centering
\setlength{\tabcolsep}{3pt}
\renewcommand{\arraystretch}{1.1}
\scriptsize
\caption{\textbf{Long-context evaluation on HELMET and RULER at context lengths up to 64K tokens.} \emph{Long-Stage1} is the intermediate checkpoint after long-context Stage~1, before STEM recovery. Per column, the best score is bold and the second-best is underlined.}
\resizebox{0.6\textwidth}{!}{%
\begin{tabular}{l|cccc|c|cccc|c}
\toprule
\textbf{Model} & \multicolumn{5}{c|}{\textbf{HELMET}} & \multicolumn{5}{c}{\textbf{RULER}} \\
\cmidrule(lr){2-6} \cmidrule(lr){7-11}
& \textbf{8K} & \textbf{16K} & \textbf{32K} & \textbf{64K} & \textbf{Avg} & \textbf{8K} & \textbf{16K} & \textbf{32K} & \textbf{64K} & \textbf{Avg} \\
\midrule
\multicolumn{11}{c}{\textit{Open-Weight Models}} \\
\midrule
Llama-3.2-3B & 42.5 & 42.6 & 38.4 & 35.0 & 39.6 & 0.0 & 0.0 & 0.0 & 0.0 & 0.0 \\
Gemma-4-E2B & 44.7 & 42.7 & 41.2 & 40.6 & 42.3 & 87.4 & 85.0 & 82.0 & 78.0 & 83.1 \\
Gemma-4-E4B & 45.0 & 43.7 & 42.9 & \underline{43.6} & 43.8 & \textbf{91.3} & \textbf{92.2} & \underline{89.6} & \underline{84.7} & \textbf{89.4} \\
Qwen3.5-2B-Base & 45.6 & 44.5 & \underline{44.5} & 42.5 & \underline{44.3} & \underline{90.9} & 87.7 & 83.0 & 83.0 & 86.1 \\
Qwen3.5-4B-Base & \textbf{52.2} & \textbf{50.9} & \textbf{48.7} & \textbf{48.8} & \textbf{50.2} & 88.1 & \underline{90.4} & \textbf{90.8} & \textbf{87.1} & \underline{89.1} \\
DeepSeek-V2-Lite & 37.9 & 35.7 & 28.3 & 27.5 & 32.4 & 76.8 & 72.0 & 58.7 & 51.1 & 64.6 \\
Moonlight-16B-A3B & 44.4 & 1.0 & 1.0 & 1.5 & 12.0 & 89.5 & 0.0 & 0.1 & 0.0 & 22.4 \\
\midrule
\multicolumn{11}{c}{\textit{Fully Open Models}} \\
\midrule
OLMoE-1B-7B & 0.4 & 1.6 & 3.0 & 3.2 & 2.1 & 0.2 & 0.0 & 0.0 & 0.0 & 0.1 \\
SmolLM3-3B-Base & 40.9 & 39.2 & 36.3 & 34.2 & 37.6 & 84.8 & 82.1 & 76.4 & 71.0 & 78.6 \\
OLMo-3-7B & \underline{47.1} & \underline{45.0} & 41.8 & 38.4 & 43.1 & 90.6 & 84.2 & 78.1 & 67.8 & 80.2 \\
\rowcolor{InstellaRowBlue}
Instella-MoE-16B-A3B-Long-Stage1 & 44.7 & 44.9 & 44.4 & 41.0 & 43.7 & 89.3 & 86.2 & 83.6 & 76.4 & 83.9 \\
\rowcolor{InstellaRowBlue}
Instella-MoE-16B-A3B-Base & 44.7 & 43.1 & 41.9 & 36.2 & 41.5 & 86.2 & 82.8 & 77.9 & 70.8 & 79.4 \\
\bottomrule
\end{tabular}%
}
\label{tab:longcontext}
\end{table*}

We evaluate long-context capability on HELMET~\citep{yen2024helmet} and RULER~\citep{hsieh2024ruler} at context lengths up to 64K tokens. HELMET evaluates retrieval-augmented generation (Natural Questions~\citep{kwiatkowski2019natural}, TriviaQA~\citep{joshi2017triviaqa}, HotpotQA~\citep{yang2018hotpotqa}), needle-in-a-haystack recall, and long-document QA (InfiniteBench~\citep{zhang2024bench}, NarrativeQA~\citep{kovcisky2018narrativeqa}).

As shown in Table~\ref{tab:longcontext}, Instella-MoE-16B-A3B-Base achieves a HELMET average of 41.5 and a RULER average of 79.4 at context lengths up to 64K tokens. On RULER, Instella-MoE is competitive with OLMo-3-7B (80.2) and edges out SmolLM3-3B (78.6), indicating that the two-stage long-context training recipe, which combines Longmino with domain-targeted data, transfers effectively to retrieval-style long-context tasks.

The intermediate Long-Stage1 checkpoint scores higher on both long-context suites, with HELMET and RULER averages of 43.7 and 83.9, respectively, so the Stage~2 STEM mixture trades part of the long-context headroom for short-context capability. We consider this an appropriate trade-off for a general-purpose base model, since the same stage lifts GSM8K from 62.8 to 81.5 and HumanEval+ from 55.0 to 65.7 (Table~\ref{tab:base_models}).

\subsection{Post-Trained Model}
\begin{table*}[t]
\centering
\setlength{\tabcolsep}{3pt}
\renewcommand{\arraystretch}{1.1}
\scriptsize
\caption{\textbf{Post-trained checkpoint results on standard benchmarks.} HE+ and LCB denote HumanEval+ and LiveCodeBench, respectively. Per column, the best score is bold and the second-best is underlined.}
\resizebox{\textwidth}{!}{%
\begin{tabular}{l|*{13}{c}}
\toprule
\textbf{Model} & \textbf{AGIEval} & \textbf{AIME24} & \textbf{AIME25} & \textbf{BBH} & \textbf{HE+} & \textbf{GPQA} & \textbf{IFEval} & \textbf{LCB} & \textbf{MBPP+} & \textbf{MATH} & \textbf{MMLU} & \textbf{AlpacaEval~2} & \textbf{Avg} \\
\midrule
\multicolumn{14}{c}{\textit{Open-Weight Models}} \\
\midrule
Gemma-4-E2B (think) & 74.99 & 48.33 & 32.08 & 72.80 & 62.32 & 46.21 & 86.88 & 46.74 & 63.10 & 86.65 & 76.49 & 40.28 & 61.41 \\
Gemma-4-E4B (think) & \underline{83.53} & 60.31 & 42.92 & \underline{76.79} & 89.63 & 50.22 & \underline{87.80} & \textbf{57.58} & \textbf{65.82} & 92.52 & \underline{83.59} & \underline{54.90} & 70.47 \\
Qwen3.5-2B & 74.25 & 6.87 & 11.46 & 70.47 & 48.35 & 50.22 & 37.52 & 8.19 & 30.21 & 48.09 & 76.10 & 29.35 & 40.92 \\
Qwen3.5-4B & \textbf{90.20} & 58.44 & 51.88 & \textbf{77.21} & 85.30 & \textbf{69.64} & 64.51 & 40.22 & 60.13 & 85.67 & \textbf{87.09} & \textbf{66.49} & 69.73 \\
DeepSeek-V2-Lite-Chat & 54.29 & 0.21 & 0.31 & 39.86 & 44.76 & 28.12 & 46.77 & 6.15 & 41.48 & 24.35 & 58.53 & 8.90 & 29.48 \\
Moonlight-16B-A3B-Instruct & 64.66 & 5.62 & 3.23 & 53.77 & 66.77 & 33.26 & 46.21 & 14.52 & 49.07 & 63.13 & 70.06 & 21.39 & 40.97 \\
\midrule
\multicolumn{14}{c}{\textit{Fully Open Models}} \\
\midrule
OLMoE-1B-7B-Instruct & 44.06 & 0.83 & 0.00 & 33.27 & 20.98 & 26.56 & 67.10 & 2.89 & 34.58 & 19.05 & 47.56 & 9.70 & 25.55 \\
SmolLM3-3B & 68.40 & 47.81 & 38.75 & 65.70 & 78.84 & 39.51 & 70.43 & 29.78 & 60.05 & 90.19 & 74.18 & 33.39 & 58.09 \\
OLMo-3-7B-Think-SFT & 78.26 & 75.21 & 64.27 & 74.80 & 88.17 & 44.42 & 81.89 & 42.99 & 63.47 & 94.67 & 76.91 & 46.66 & 69.31 \\
OLMo-3-7B-Think-DPO & 81.62 & 75.31 & 66.46 & 72.09 & \textbf{90.00} & 44.87 & 77.26 & 45.93 & 64.47 & \textbf{95.48} & 79.09 & 45.19 & 69.81 \\
OLMo-3-7B-Think & 81.50 & 75.00 & 69.17 & 69.67 & \underline{89.88} & 45.54 & \textbf{90.39} & 51.94 & \underline{64.84} & \underline{95.24} & 79.07 & 51.39 & 71.97 \\
\rowcolor{InstellaRowBlue} Instella-MoE-16B-A3B-SFT & 81.33 & \textbf{77.50} & 71.35 & 74.34 & 86.77 & 59.38 & 81.51 & 48.74 & 61.64 & 94.67 & 80.82 & 40.92 & 71.58 \\
\rowcolor{InstellaRowBlue} Instella-MoE-16B-A3B-DPO & 81.51 & \underline{77.40} & \underline{73.23} & 74.77 & 86.52 & \underline{62.72} & 77.08 & 53.39 & 61.98 & 94.92 & 81.74 & 46.80 & \underline{72.67} \\
\rowcolor{InstellaRowBlue}
Instella-MoE-16B-A3B-Think & 82.50 & 76.90 & \textbf{73.40} & 74.80 & 87.90 & 61.20 & 83.70 & \underline{54.30} & 62.00 & 94.80 & 81.30 & 45.81 & \textbf{73.22} \\
\bottomrule
\end{tabular}%
}
\label{tab:posttrain}
\end{table*}

We evaluate our post-trained checkpoints on instruction-following, reasoning, and chat benchmarks, including AGIEval~\citep{zhong2024agieval}, AIME24/25, BBH~\citep{suzgun2022challengingbigbenchtaskschainofthought}, GPQA~\citep{rein2023gpqagraduatelevelgoogleproofqa}, HumanEval+ and MBPP+~\citep{evalplus}, IFEval~\citep{zhou2023instructionfollowingevaluationlargelanguage}, LiveCodeBench (LCB)~\citep{jain2025livecodebench}, MATH~\citep{hendrycksmath2021}, MMLU~\citep{hendrycks2021measuringmassivemultitasklanguage}, and AlpacaEval~2~\citep{dubois2025lengthcontrolledalpacaevalsimpleway}.

As shown in Table~\ref{tab:posttrain} and Figure~\ref{fig:cost_performance} (right), SFT equips the base model with instruction-following and chain-of-thought capabilities using the Dolci-Think and Nemotron mixtures, achieving an average score of 71.6. Our MoE-specific router stabilization recipe enables DPO to further improve helpfulness and response quality, raising the average to \textbf{72.7} while preserving reasoning performance. Our final Think checkpoint reaches an average of \textbf{73.2}, the strongest fully open post-trained result in our comparison at 2.8B active parameters.

\subsection{Ablation Studies}

\subsubsection{Architecture Ablation}
\begin{table*}[t]
\centering
\setlength{\tabcolsep}{3pt}
\renewcommand{\arraystretch}{1.1}
\scriptsize
\caption{\textbf{Architecture ablation.} Effect of the Gated MLA output gate and FarSkip-Collective connectivity, measured after 200B tokens (Nemotron-CC, 48K steps). \emph{Vanilla} is plain MLA with standard expert-parallel connectivity. Gated MLA drives the accuracy improvement, and adding FarSkip-Collective maintains average accuracy similar to that of Gated MLA while enabling the communication-overlap efficiency gains. HE+ denotes HumanEval+. Per column, the best score is bold and the second-best is underlined; all results are from a single run.}
\resizebox{\textwidth}{!}{%
\begin{tabular}{l|*{15}{c}|c}
\toprule
\textbf{Model} & \textbf{ARC-C} & \textbf{ARC-E} & \textbf{BoolQ} & \textbf{HSwag} & \textbf{MMLU} & \textbf{OBQA} & \textbf{PIQA} & \textbf{SciQ} & \textbf{WG} & \textbf{BBH} & \textbf{GSM8K} & \textbf{MATH} & \textbf{GPQA} & \textbf{HE+} & \textbf{MBPP+} & \textbf{Avg} \\
\midrule
Vanilla (MLA) & 43.34 & \textbf{70.79} & 64.50 & \textbf{69.98} & 39.02 & 39.40 & \underline{78.45} & \textbf{92.40} & \textbf{63.77} & \textbf{38.26} & \textbf{32.22} & \textbf{24.54} & 23.44 & 21.95 & 45.77 & 49.86 \\
Gated MLA & \underline{43.94} & \underline{70.33} & \underline{65.72} & 69.23 & \textbf{43.28} & \textbf{41.00} & 78.24 & 90.50 & \underline{62.35} & 35.37 & 30.93 & \underline{23.26} & \underline{25.22} & \textbf{26.83} & \underline{48.68} & \underline{50.33} \\
\rowcolor{InstellaRowBlue}
Gated MLA + FarSkip & \textbf{44.62} & 69.91 & \textbf{68.04} & \underline{69.47} & \underline{41.86} & \underline{40.60} & \textbf{79.00} & \underline{90.90} & 62.19 & \underline{35.54} & \underline{31.39} & 22.34 & \textbf{26.79} & \underline{23.78} & \textbf{49.21} & \textbf{50.38} \\
\bottomrule
\end{tabular}%
}
\label{tab:arch_ablation}
\end{table*}

To evaluate the contribution of the Gated MLA output gate and FarSkip-Collective connectivity, we run a controlled architectural ablation at a 200B-token budget on the Nemotron-CC~\citep{su2025nemotroncc} dataset, using Vanilla MLA with standard expert-parallel connectivity as the baseline. We use LM-Eval-Harness~\citep{lm-eval-harness} for evaluation. As shown in Table~\ref{tab:arch_ablation}, Gated MLA improves the average from 49.86 to 50.33 ($+0.47$), with the clearest gains on code (HumanEval+ $+4.9$, MBPP+ $+2.9$), MMLU ($+4.3$), and GPQA ($+1.8$). Incorporating FarSkip-Collective connectivity achieves a similar average accuracy of 50.38 while delivering the communication-overlap efficiency gains reported in Figure~\ref{fig:efficiency}.

\subsubsection{Reinforcement Learning Ablation}
\begin{table}[t]
\centering
\setlength{\tabcolsep}{3pt}
\renewcommand{\arraystretch}{1.1}
\scriptsize
\caption{\textbf{RL ablation results.} IF-RL improves instruction following but regresses on math and reasoning benchmarks; MOPD recovers most of the instruction-following gains while preserving DPO-level performance elsewhere. HE+ and LCB denote HumanEval+ and LiveCodeBench, respectively. Per column, the best score is bold and the second-best is underlined.}
\resizebox{0.9\columnwidth}{!}{%
\begin{tabular}{l|ccccccccccc|c}
\toprule
\textbf{Stage} & \textbf{AGIEval} & \textbf{AIME24} & \textbf{AIME25} & \textbf{BBH} & \textbf{HE+} & \textbf{GPQA} & \textbf{IFEval} & \textbf{LCB} & \textbf{MBPP+} & \textbf{MATH} & \textbf{MMLU} & \textbf{Avg} \\
\midrule
DPO & \underline{81.5} & \textbf{77.4} & \underline{73.2} & \underline{74.8} & \underline{86.5} & \textbf{62.7} & 77.1 & \underline{53.4} & \underline{62.0} & \textbf{94.9} & \textbf{81.7} & \underline{75.0} \\
IF-RL & 81.3 & 74.3 & 65.2 & \textbf{75.0} & 85.5 & 60.9 & \textbf{84.1} & 51.5 & \textbf{63.2} & 92.1 & \underline{81.3} & 74.0 \\
\rowcolor{InstellaRowBlue}
MOPD & \textbf{82.5} & \underline{76.9} & \textbf{73.4} & \underline{74.8} & \textbf{87.9} & \underline{61.2} & \underline{83.7} & \textbf{54.3} & \underline{62.0} & \underline{94.8} & \underline{81.3} & \textbf{75.7} \\
\bottomrule
\end{tabular}%
}
\label{tab:rl_ablation}
\end{table}

We obtain our final Think checkpoint by applying a two-stage RL recipe to the DPO checkpoint (Section~\ref{sec:rl}). Table~\ref{tab:rl_ablation} summarizes the ablation. IF-specialized RL alone improves IFEval from 77.1 to 84.1 but regresses on AIME24/25 and GPQA. Multi-Teacher On-Policy Distillation (MOPD) recovers most of the IF expert's instruction-following gain (IFEval 83.7) while maintaining DPO-level performance on math, code, MMLU, and AGIEval, yielding the highest 11-benchmark ablation average of \textbf{75.7}. This indicates that domain-routed on-policy distillation is an effective strategy for adding targeted RL capabilities to MoE models without catastrophic forgetting in other domains.

\subsection{Training Efficiency}
\label{sec:efficiency}
\begin{figure}[t]
    \centering
    \includegraphics[width=\linewidth]{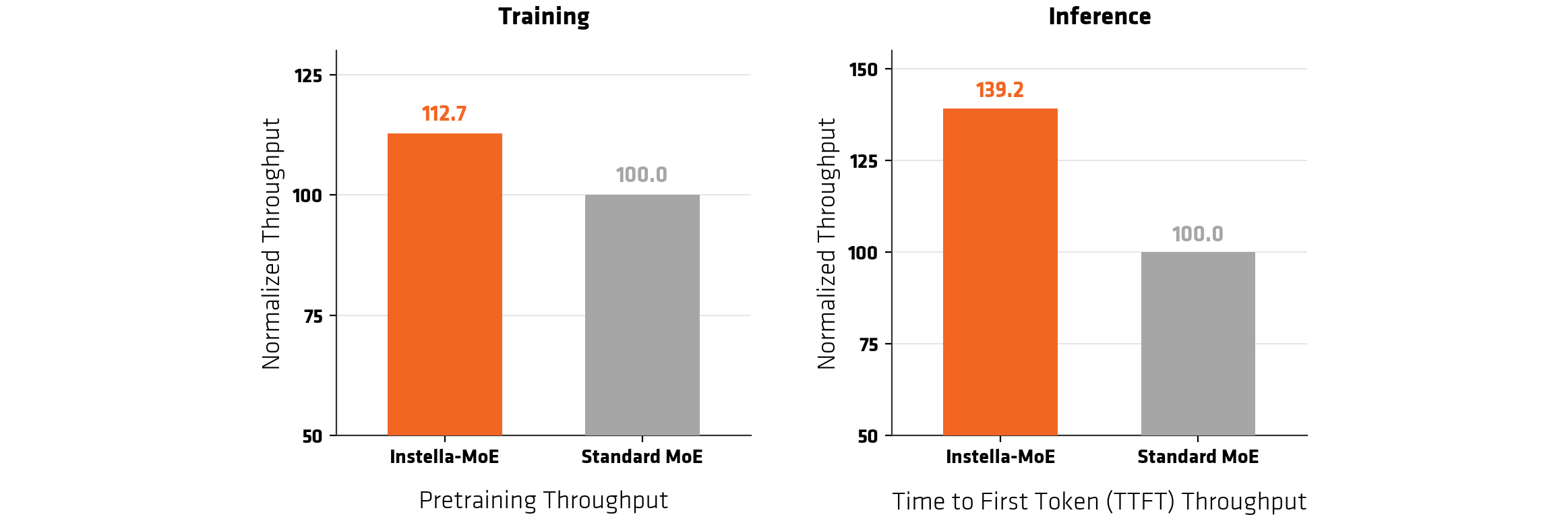}
    \caption{\textbf{Instella-MoE-16B-A3B training and inference throughput.} \emph{Left:} pre-training throughput, normalized to a standard MoE expert-parallel training baseline (12.7\% improvement). \emph{Right:} inference time-to-first-token (TTFT) throughput under expert parallelism (39.2\% improvement).}
    \label{fig:efficiency}
\end{figure}

Beyond benchmark accuracy, Instella-MoE is designed for efficient training and serving on AMD hardware. FarSkip-Collective overlap yields a 12.7\% pre-training throughput improvement over the standard expert-parallel baseline architecture. At inference, SGLang serving with expert parallelism raises time-to-first-token (TTFT) throughput by 39.2\% (Figure~\ref{fig:efficiency}) by overlapping model communication with computation.These system-level optimizations complement the parameter efficiency of the MoE architecture by removing the significant communication overhead of MoEs and harnessing the efficiency of the sparse architecture with 2.8B active parameters per token.

\section{Conclusion}
We present Instella-MoE-16B-A3B, a fully open Mixture-of-Experts language model trained entirely on openly available data, using open-source code and AMD Instinct GPU infrastructure. With 16 billion total parameters and 2.8 billion active parameters per token, Instella-MoE combines the efficiency of sparse MoE architectures with architectural and system-level innovations---Gated Multi-head Latent Attention and FarSkip-Collective connectivity---to achieve performance competitive with both fully open and open-weight baselines while keeping per-token compute close to that of a dense 3B-class model.

The Instella-MoE release spans the complete model flow: pre-training, mid-training, long-context extension, supervised fine-tuning, direct preference optimization, and reinforcement learning. Instella-MoE-16B-A3B-Base achieves state-of-the-art performance among fully open models in our evaluation, with an average score of 76.7 across standard pre-training benchmarks, and our final Think checkpoint delivers the strongest overall post-trained results in our comparison. Alongside model weights from every training stage, we release training configurations, data mixtures, evaluation protocols, and training code built on the Primus and Miles frameworks.

Instella-MoE demonstrates that transparency and strong performance can coexist in modern sparse MoE systems. By open-sourcing a complete pipeline for training an MoE model from scratch on AMD hardware, we aim to support reproducible research, community-driven innovation, and further advances in efficient open language modeling.

\section*{Acknowledgements}
\addcontentsline{toc}{section}{Acknowledgements}

We are deeply grateful to the LLM360 team and the Miles team for their invaluable support throughout the development of our model. We dedicate this work to the memory of Zicheng Liu, whose vision and leadership continue to inspire our team.

\bibliography{main}
\bibliographystyle{colm2024_conference}

\newpage
\appendix
\section{Overlapped Implementation of Instella-MoE}
 \label{app:overlapped_implementation}

 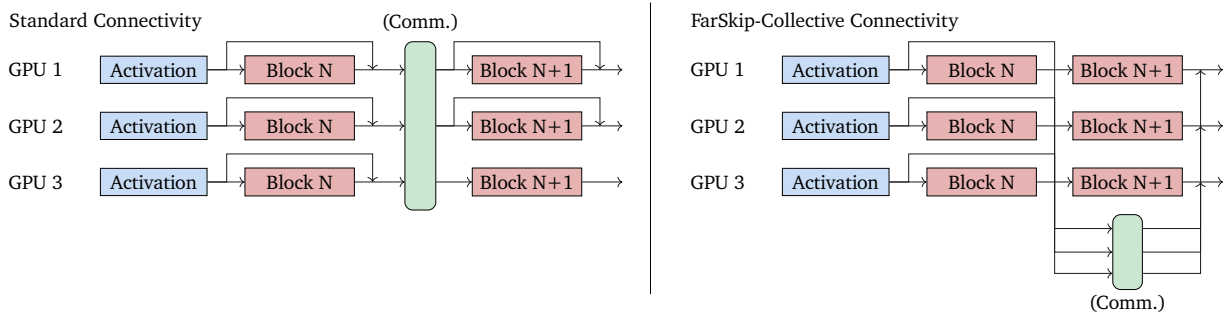
\begin{figure}[t]
    \centering
    \resizebox{\linewidth}{!}{\definecolor{ink}{HTML}{1C1C1E}
\definecolor{attnblue}{HTML}{C8DCF5}
\definecolor{mlared}{HTML}{E6B3B2}
\definecolor{moegreen}{HTML}{CBE7CF}

\begin{tikzpicture}[
    act/.style={draw, fill=attnblue, minimum width=1.9cm},
    blk/.style={draw, fill=mlared, minimum width=1.95cm},
    comm/.style={draw, shape=rectangle, rounded corners, fill=moegreen},
    arr/.style={->, ink}]

  \node[draw=none] at (0,3) {GPU 3};
  \node[draw=none] at (0,4) {GPU 2};
  \node[draw=none] at (0,5) {GPU 1};
  \node[act] (node5) at (2.1,3) {Activation};
  \node[act] (node3) at (2.1,4) {Activation};
  \node[act] (node1) at (2.1,5) {Activation};
  \node[blk] (node6) at (4.7,3) {Block N};
  \node[blk] (node4) at (4.7,4) {Block N};
  \node[blk] (node2) at (4.7,5) {Block N};
  \node[comm, minimum width=0.55cm, minimum height=3cm] (node7) at (6.85,4) {};
  \node[blk] (node8)  at (8.75,3) {Block N+1};
  \node[blk] (node9)  at (8.75,4) {Block N+1};
  \node[blk] (node10) at (8.75,5) {Block N+1};

  \draw[arr] (node1.east) -- (3.4,5) -- (3.4,5.5) -- (6.0,5.5) -- (6.0,5);
  \draw[arr] (node1.east) -- (node2.west);
  \draw[arr] (node3.east) -- (3.4,4) -- (3.4,4.5) -- (6.0,4.5) -- (6.0,4);
  \draw[arr] (node3.east) -- (node4.west);
  \draw[arr] (node5.east) -- (3.4,3) -- (3.4,3.5) -- (6.0,3.5) -- (6.0,3);
  \draw[arr] (node5.east) -- (node6.west);
  \draw[arr] (node2.east) -- (6.57,5);
  \draw[arr] (node4.east) -- (node7.west);
  \draw[arr] (node6.east) -- (6.57,3);
  \draw[arr] (7.13,5) -- (7.45,5) -- (7.45,5.5) -- (10.05,5.5) -- (10.05,5);
  \draw[arr] (7.13,5) -- (node10.west);
  \draw[arr] (node7.east) -- (7.45,4) -- (7.45,4.5) -- (10.05,4.5) -- (10.05,4);
  \draw[arr] (node7.east) -- (node9.west);
  \draw[arr] (7.13,3) -- (node8.west);
  \draw[arr] (node10.east) -- (10.45,5);
  \draw[arr] (node9.east) -- (10.45,4);
  \draw[arr] (node8.east) -- (10.45,3);
  \node[draw=none] at (6.85,5.85) {(Comm.)};
  \node[draw=none, anchor=west] at (-0.6,5.85) {Standard Connectivity};

  \draw (10.95,6.2) -- (10.95,1);

  \node[draw=none] at (12.15,3) {GPU 3};
  \node[draw=none] at (12.15,4) {GPU 2};
  \node[draw=none] at (12.15,5) {GPU 1};
  \node[act] (node11) at (14.25,3) {Activation};
  \node[act] (node12) at (14.25,4) {Activation};
  \node[act] (node13) at (14.25,5) {Activation};
  \node[blk] (node14) at (16.85,3) {Block N};
  \node[blk] (node15) at (16.85,4) {Block N};
  \node[blk] (node16) at (16.85,5) {Block N};
  \node[comm, minimum width=15pt, minimum height=38pt] (node17) at (19.45,1.75) {};
  \node[blk] (node18) at (19.45,3) {Block N+1};
  \node[blk] (node19) at (19.45,4) {Block N+1};
  \node[blk] (node20) at (19.45,5) {Block N+1};

  \draw[arr] (node13.east) -- (15.55,5) -- (15.55,5.5) -- (18.15,5.5) -- (18.15,2.17) -- (19.18,2.17);
  \draw[arr] (node13.east) -- (node16.west);
  \draw[arr] (node12.east) -- (15.55,4) -- (15.55,4.5) -- (18.15,4.5) -- (18.15,1.75) -- (node17.west);
  \draw[arr] (node12.east) -- (node15.west);
  \draw[arr] (node11.east) -- (15.55,3) -- (15.55,3.5) -- (18.15,3.5) -- (18.15,1.37) -- (19.18,1.37);
  \draw[arr] (node11.east) -- (node14.west);
  \draw[arr] (node16.east) -- (node20.west);
  \draw[arr] (node15.east) -- (node19.west);
  \draw[arr] (node14.east) -- (node18.west);
  \draw[arr] (node20.east) -- (21.15,5);
  \draw[arr] (node19.east) -- (21.15,4);
  \draw[arr] (node18.east) -- (21.15,3);
  \draw[arr] (19.72,2.17) -- (20.75,2.17) -- (20.75,5);
  \draw[arr] (node17.east) -- (20.75,1.75) -- (20.75,4);
  \draw[arr] (19.72,1.37) -- (20.75,1.37) -- (20.75,3);
  \node[draw=none] at (19.45,0.83) {(Comm.)};
  \node[draw=none, anchor=west] at (11.55,5.85) {FarSkip-Collective Connectivity};
\end{tikzpicture}}
    \caption{\textbf{FarSkip-Collective connectivity}~\citep{dukler2026farskipcollectiveunhobblingblockingcommunication}. FarSkip-Collective modifies the connectivity between sub-blocks to avoid waiting for communication collectives. Computation continues with available activations, partial (e.g., Block N output) or outdated (e.g., Activation).}
    \label{fig:farskip_connectivity}
\end{figure}

Instella-MoE employs FarSkip-Collective connectivity, which overlaps communication with computation by passing outdated and partial activations into MoE and attention layers (Figure~\ref{fig:farskip_connectivity}), reducing communication bubbles while maintaining model accuracy. To explicitly achieve communication overlap during Instella-MoE training with FarSkip-Collective, we develop a custom model implementation that modularly enforces hardware overlap while avoiding custom kernels for portability. We go over our implementation techniques below.

 On GPU hardware, communication-computation overlap ensures computation continues throughout the workload while communication is running simultaneously. This improves utilization and relies on sharing CUs between communication and computation kernels during runtime.
 The standard approach to implementing overlap is relying on the asynchronous launching behavior of kernels via \texttt{torch.distributed}'s \texttt{async\_op=True} flag that will yield back after queuing the kernel. This is coupled with \texttt{.wait()} barriers that are used for safe access of the communicated tensors.
 We use this approach during the forward pass as we explicitly order the model operations to maximize overlap. Nonetheless, during the backward pass, using \texttt{torch.autograd} makes explicit operator ordering and \texttt{.wait()} calls inaccessible.
 To resolve this we use the approach proposed by \citet{dukler2026farskipcollectiveunhobblingblockingcommunication} (see Appendix~C therein), developing race-safe asynchronous backward communication operations and combining them with sequence number re-prioritization using the Sequence Number API.\footnote{\url{https://github.com/pytorch/pytorch/blob/main/aten/src/ATen/SequenceNumber.h}}
Benchmarking our overlapped implementation during pre-training against the standard connectivity variant, we achieve a 12.7\% training speed-up (Figure~\ref{fig:efficiency}).

For inference, we implement communication-overlapped expert-parallel generation with support for both tensor-parallel attention and data-parallel attention to support Instella-MoE's Gated-MLA-based attention. In addition to overlapping, by aggregating communication we are able to sum and communicate combined tensors, which allows us to further reduce the total communication volume.
Using our implementation we raise time-to-first-token (TTFT) throughput by 39.2\% by overlapping communication when serving Instella-MoE using 8$\times$MI325X GPUs (Figure~\ref{fig:efficiency}). The exact overlapped implementation code is included and documented in our release \href{https://github.com/AMD-AGI/Instella-MoE}{\texttt{github.com/AMD-AGI/Instella-MoE}}.

\section{SFT Data Curation Details}
\label{app:sft_curation}

We provide additional details on the feedback-driven final-stage data curation
introduced in the main text. We first present representative teacher error
analyses and the retrieval queries derived from them, then describe how these
queries are converted into selected training examples, and finally specify the
underlying data splits and selection budget.

\textbf{Error analyses and policies:} each seed problem that the student answers
incorrectly is passed to the judge model, which returns a structured error analysis whose
fields record the failure mode, the skills the student lacks, and a description of
the training data that would address the gap. Tables~\ref{tab:teacher_analyses_math}
and~\ref{tab:teacher_analyses_code} report two such analyses, one for a
mathematics problem and one for a coding problem, with the student's
response summarized rather than reproduced verbatim. Within each domain, the
reflection model aggregates these per-problem analyses into a \emph{policy}: a set
of retrieval queries whose weights are normalized to one and collectively define the target
distribution for selection. Table~\ref{tab:policy_queries} shows five queries
sampled at random from each targeted-domain policy, with those queries sorted by weight.

\begin{table}[htbp]
\centering
\small
\caption{\textbf{Teacher error analysis (mathematics).} Structured analysis produced by the judge model for a representative reasoning failure. The analysis is presented here under the consolidated fields \emph{domain}, \emph{complexity}, \emph{error}, \emph{skills}, \emph{would help}, and \emph{summary}; the problem statement is shown only for context.}
\setlength{\tabcolsep}{5pt}
\renewcommand{\arraystretch}{1.1}
\begin{tabular}{@{}>{\bfseries}l L{0.80\linewidth}@{}}
\toprule
\rowcolor{PromptBg}
\multicolumn{2}{@{}L{0.97\linewidth}@{}}{\textbf{Problem (reference answer 503).} The probability that a set of three distinct vertices chosen at random from among the vertices of a regular $n$-gon determine an obtuse triangle is $\frac{93}{125}$. Find the sum of all possible values of $n$.} \\
\midrule
Domain & math $\rightarrow$ combinatorics and geometry $\rightarrow$ Determining the probability that three randomly chosen vertices of a regular $n$-gon form an obtuse triangle by analyzing arc lengths and using symmetry, then solving for $n$ given a specific probability value \\
\addlinespace[3pt]
Complexity & Multi-step: requires understanding geometric properties of cyclic polygons, distinguishing between acute, right, and obtuse triangles based on arc measures, deriving general formulas for the number of obtuse triangles in terms of $n$ for both even and odd cases, setting up and solving rational equations, and verifying solutions against constraints. \\
\addlinespace[3pt]
Error & Incorrect approach and incomplete reasoning: focused on a specific small case (hexagon) instead of generalizing, made counting errors, misapplied angle formulas, failed to derive the correct general formula, and did not solve for $n$ \\
\addlinespace[3pt]
Skills & The student lacks the ability to generalize from a specific example to a general formula, does not understand the relationship between central arcs and inscribed angles in a circle, fails to recognize that only one angle in a triangle can be obtuse and that it corresponds to the largest arc opposite it, and does not know how to count triangles lying on a semicircle or subtract right triangles appropriately. Additionally, the student does not know how to set up and solve equations involving binomial coefficients and rational expressions to find integer solutions. \\
\addlinespace[3pt]
Would help & Problems involving probability of triangle types in regular polygons with step-by-step solutions that derive general formulas using symmetry and arc-length arguments, especially those that distinguish between even and odd $n$, explain the connection between inscribed angles and arcs, and show how to count configurations on semicircles. Examples should include worked derivations of the number of obtuse triangles as a function of $n$, followed by solving equations like $3(n-3)/(4(n-2)) = 93/125$. \\
\addlinespace[3pt]
Summary & The student incorrectly focused on a specific case ($n=6$) and attempted to manually count obtuse triangles by examining arcs, but failed to generalize the method to arbitrary $n$. They made errors in counting pairs and misapplied geometric principles, such as incorrectly relating arc length to angle measure. The reasoning never led to a general formula for the number of obtuse triangles, nor did it result in solving the equation for $n$. The correct approach requires deriving separate formulas for odd and even $n$ based on semicircle containment and right triangle counts, then solving the resulting rational equations. The student's work is incomplete, inconsistent, and does not reach the correct values of $n$ or their sum. \\
\bottomrule
\end{tabular}
\label{tab:teacher_analyses_math}
\end{table}

\begin{table}[htbp]
\centering
\small
\caption{\textbf{Teacher error analysis (coding).} Structured analysis produced by the judge model for a representative reasoning failure, presented under the same consolidated fields as Table~\ref{tab:teacher_analyses_math}; the problem statement is shown only for context.}
\setlength{\tabcolsep}{5pt}
\renewcommand{\arraystretch}{1.1}
\begin{tabular}{@{}>{\bfseries}l L{0.80\linewidth}@{}}
\toprule
\rowcolor{PromptBg}
\multicolumn{2}{@{}L{0.97\linewidth}@{}}{\textbf{Problem.} A dog starts at $x=0$ and runs right at unit speed past $n$ bowls at $x=1,\dots,n$; bowl $i$ becomes edible only at time $t_i$. On reaching a hot bowl the dog may wait until $t_i$ (delaying every later bowl) or skip it and continue. Maximize the number of bowls eaten within $T$ seconds.} \\
\midrule
Domain & algorithmic problem solving $\rightarrow$ dynamic programming with greedy optimization and binary search $\rightarrow$ strategic decision-making in a time-constrained sequential process where the choice to skip a valid option may enable a better overall outcome, requiring a non-greedy, globally optimal strategy rather than a local greedy one \\
\addlinespace[3pt]
Complexity & multi-step with backtracking and optimization: requires recognizing that a greedy approach of eating every valid bowl immediately is suboptimal, understanding the trade-off between waiting at a bowl and skipping it to reach more bowls later, and implementing a binary search with heap-based feasibility checking to efficiently evaluate the maximum number of bowls that can be eaten under constraints \\
\addlinespace[3pt]
Error & incorrect greedy assumption and failure to model strategic skipping decisions \\
\addlinespace[3pt]
Skills & the student lacks the ability to recognize that optimal solutions in sequential decision problems with time constraints may require skipping seemingly beneficial options to achieve a better long-term outcome. They also lack familiarity with advanced techniques such as binary search over answers combined with heap-based maintenance of critical values (like $t_j - j$) to simulate optimal skipping strategies. Additionally, they fail to understand how the timing of arrival and cooling interacts with the decision to wait or skip. \\
\addlinespace[3pt]
Would help & problems involving optimal scheduling with trade-offs between immediate action and future opportunities, especially those requiring binary search on answer combined with heap or priority queue data structures to maintain state for feasibility checks. Examples should include scenarios where skipping a valid option enables a better overall result, such as `minimum time to complete tasks with dependencies and cooldowns' or `maximize number of jobs with deadlines and processing times'. Worked solutions should explicitly compare greedy vs.\ optimal strategies and demonstrate why greedy fails in certain cases. \\
\addlinespace[3pt]
Summary & The student incorrectly assumes that eating every bowl for which $\max(i+1, t_i) < T$ is optimal, failing to recognize that skipping an early valid bowl might allow the dog to eat more bowls later by avoiding delays. This greedy approach ignores the strategic trade-off between waiting at a bowl (which delays the dog) and skipping it (which may enable reaching more bowls within the time limit). The correct solution uses binary search on the number of bowls and a heap to track the $k-1$ smallest values of $(t_j - j)$ in the prefix, which helps determine whether skipping certain bowls leads to a better overall outcome. The student's reasoning is limited to local validity checks and does not account for the global impact of skipping decisions, leading to a fundamentally flawed strategy. \\
\bottomrule
\end{tabular}
\label{tab:teacher_analyses_code}
\end{table}

\begin{table}[htbp]
\centering
\small
\caption{\textbf{Sample retrieval queries.} Five queries drawn at random from each targeted-domain policy.}
\setlength{\tabcolsep}{5pt}
\renewcommand{\arraystretch}{1.1}
\begin{tabular}{@{}lrL{0.74\linewidth}@{}}
\toprule
\textbf{Domain} & \textbf{Weight} & \textbf{Query} \\
\midrule
\textbf{Math} & 0.017 & Coordinate geometry problems involving polygons with area equality constraints, fixed side alignments on triangle edges, or intersection conditions, requiring use of shoelace formula, vector dot products, and solving systems of equations derived from geometric constraints \\
\addlinespace[3pt]
 & 0.009 & Complex number geometry problems involving regular polygons inscribed in circles, roots of unity, rotational symmetries, and magnitude calculations of vector sums, requiring trigonometric identities and algebraic manipulation of complex expressions \\
\addlinespace[3pt]
 & 0.009 & Prime factorization analysis and divisor counting, including odd and even divisor separation and unordered factorization counting \\
\addlinespace[3pt]
 & 0.007 & Digit constraint problems including 9's complement methods, counting numbers with specific digit properties, avoiding zero digits, and optimization over digit assignments \\
\addlinespace[3pt]
 & 0.005 & Diophantine equation problems involving counting positive integer solutions to linear equations with bounds, constraints on digit sums, and modular restrictions \\
\midrule
\textbf{Code} & 0.020 & Graph algorithms focusing on BFS for shortest paths in state spaces, DFS for connectivity and component analysis, and Dijkstra's algorithm for weighted shortest paths. Include problems with mathematical state spaces involving rational numbers, grid-based movement, and tree structures. Emphasize proper implementation of visited sets, queue management, and path reconstruction. \\
\addlinespace[3pt]
 & 0.018 & Strict input/output format compliance problems emphasizing precise input handling (preserving whitespace and avoiding premature \texttt{.strip()}), exact output string templates, JSON structure validation, and adherence to specific message formats. Include tasks where output formatting is as critical as algorithmic correctness, such as exact word counts, specific capitalization rules, and structured data output. \\
\addlinespace[3pt]
 & 0.005 & Spatial graph connectivity problems defined on geometric grids where adjacency depends on Manhattan distance, same-row or same-column proximity, or visibility lines. Include problems requiring sorting of coordinates by x or y values and connecting only spatially adjacent points, avoiding abstract bipartite graph misrepresentations. \\
\addlinespace[3pt]
 & 0.003 & Prolog and logic programming: predicate chaining, transformation pipelines (named\_to\_op\_expr, prefix token conversion), compound term handling, and constraint logic programming without direct recursion. \\
\addlinespace[3pt]
 & 0.003 & Web application development using frameworks like Streamlit including API integration, HTTP requests, JSON response parsing, and interactive UI components. Include data persistence, state management across sessions, and responsive layout implementation. \\
\bottomrule
\end{tabular}
\label{tab:policy_queries}
\end{table}

\textbf{From queries to selected examples:} each targeted-domain policy is a list
of retrieval queries with non-negative weights that sum to one. Selection uses a
two-level allocation of the $512$K budget $B$. First, $B$ is split across
domains in the same proportions as the base mixture, giving a per-domain budget
$B_{\mathrm{dom}}$. Within a targeted domain, a query with weight $w_i$ retrieves
$\lfloor B_{\mathrm{dom}}\, w_i (1-\alpha) \rfloor$ nearest neighbors from the pool
by cosine similarity in a shared embedding space. Because the weights sum to one,
the retrieved examples account for approximately a $(1-\alpha)$ share of the domain budget after integer rounding; the
remaining budget is filled with random same-domain examples. This random
component preserves coverage and prevents the mixture from collapsing onto the
retrieved neighborhoods. We use $\alpha = 0.5$. The non-targeted \emph{other}
domain has no policy, so its entire budget is filled by random sampling within
that domain.

\textbf{Data splits and budget:} the curation operates over four disjoint sets
that mirror the stages of the pipeline. Diagnosis and query generation run on a
held-out \emph{seed set} of problems paired with reference answers: for
mathematics we use AIME problems through $2022$, and for coding we use problems
drawn from the Nemotron pool. To measure the effect of curation without leakage, a
separate \emph{scoring set} is reserved for evaluation and never enters selection.
Retrieval then draws from the full instruction-tuning \emph{training pool}. We
remove any content that overlaps with either the seed set or the scoring set so
that neither diagnostic nor evaluation problems can be reselected, and we drop
any example longer than $32$K tokens before embedding. The examples returned by retrieval, together with the random
same-domain fill described above, form the \emph{curated set}: the final $512$K
mixture used for the last SFT phase, which contains the same number of samples per
domain as the base mixture.

\textbf{Curation prompts:} for reproducibility, we reproduce the prompts used by
the pipeline verbatim. The judge model runs the correctness-verification prompt
(Table~\ref{tab:prompt_judge}) and, on incorrect responses, one of two
error-analysis prompts---a general one (Table~\ref{tab:prompt_feedback}) and a
coding-specific one (Table~\ref{tab:prompt_feedback_coding}). The reflection model
runs the policy prompt (Table~\ref{tab:prompt_reflection}) that turns the
aggregated error analyses into weighted retrieval queries.

\begin{table}[htbp]
\caption{\textbf{Judge prompt.} System prompt used by the LLM-as-judge to verify a student response against the reference answer.}
\label{tab:prompt_judge}
\begin{promptbox}
You are a strict answer verifier. You will see a question, a reference (gold) solution, and a student's solution. Determine whether the student arrived at the correct final answer or output.\newline
\newline
The task may be math, coding, science, instruction following, creative writing, or anything else. Judge correctness based on the task type:\newline
~~- Math/science: equivalent final answers count as correct (0.5 = 1/2, etc.)\newline
~~- Code: output must be functionally equivalent (ignore whitespace, variable names)\newline
~~- Instruction following: the student must satisfy the stated requirements\newline
~~- Open-ended: use the reference as a guide; accept reasonable equivalents\newline
\newline
Respond with EXACTLY one JSON object on a single line:\newline
\{"correct": true\}~~or~~\{"correct": false, "reason": "\textless{}brief explanation\textgreater{}"\}
\end{promptbox}
\end{table}

\begin{table}[htbp]
\caption{\textbf{Error-analysis prompt (general).} System prompt used by the judge model to produce a structured error analysis for an incorrect response.}
\label{tab:prompt_feedback}
\begin{promptbox}
You are analyzing why a student AI model produced a wrong answer. The task could be math, coding, science, instruction following, creative writing, or anything else. Given the question, reference solution, and the student's incorrect output, provide detailed structured feedback. Be verbose and descriptive in every field --- write in natural language, not just labels.\newline
\newline
Respond with EXACTLY one JSON object:\newline
\{\newline
~~"domain": "\textless{}broad area: math, code, science, instruction\_following, creative, general, etc.\textgreater{}",\newline
~~"sub\_domain": "\textless{}specific sub-field within the domain, e.g. 'combinatorics and counting', 'dynamic programming', 'thermodynamics', 'output format compliance'\textgreater{}",\newline
~~"concept": "\textless{}describe the specific concept or skill being tested in plain English, e.g. 'applying the inclusion-exclusion principle to count overlapping sets', 'handling recursive base cases in tree traversal', 'following multi-part formatting instructions precisely'\textgreater{}",\newline
~~"reasoning\_complexity": "\textless{}describe what level of reasoning this task requires and why, e.g. 'multi-step: requires setting up equations from word problem, solving the system, then verifying constraints', 'single-step: direct formula application', 'multi-step with backtracking: requires trying cases and eliminating invalid ones'\textgreater{}",\newline
~~"error\_type": "\textless{}describe what went wrong in a short phrase, e.g. 'confused area formula with circumference formula', 'off-by-one in loop boundary and wrong conditional ordering', 'ignored explicit bullet point and verb-first requirements'\textgreater{}",\newline
~~"skills\_lacking": "\textless{}describe in detail what knowledge or abilities the student is missing, e.g. 'the student needs a firmer grasp of when to apply $\pi r^2$ vs $2\pi r$, and more generally how to distinguish between area and perimeter formulas across shapes'\textgreater{}",\newline
~~"what\_would\_help": "\textless{}describe what kind of training data would help the student get better at this, written as a description suitable for retrieving similar examples from a large pool, e.g. 'geometry problems that require choosing between area, perimeter, and volume formulas for circles, rectangles, and spheres, with explicit worked solutions showing formula selection reasoning'\textgreater{}",\newline
~~"summary": "\textless{}2-4 sentence detailed explanation of what went wrong, why the student's approach failed, and what the correct approach would have been\textgreater{}"\newline
\}
\end{promptbox}
\end{table}

\begin{table}[htbp]
\caption{\textbf{Error-analysis prompt (coding).} Coding-specific variant of the error-analysis prompt, used by the judge model for problems in the code domain.}
\label{tab:prompt_feedback_coding}
\begin{promptbox}
You are analyzing why a student AI model produced a wrong answer on a CODING task. All tasks in this set require writing code to solve --- including scripts, automation, API usage, data processing, file manipulation, web development, and system administration tasks. Even if a task looks like "instruction following", the solution requires functional code.\newline
\newline
Given the question, reference solution, and the student's incorrect output, provide detailed structured feedback focused on the coding skills needed. Be verbose and descriptive in every field --- write in natural language, not labels.\newline
\newline
Respond with EXACTLY one JSON object:\newline
\{\newline
~~"domain": "code",\newline
~~"sub\_domain": "\textless{}specific coding sub-field, e.g. 'string processing', 'file I/O and scripting', 'API integration', 'data parsing and transformation', 'web scraping', 'command-line tool usage', 'database queries', 'text encoding and Unicode handling'\textgreater{}",\newline
~~"concept": "\textless{}describe the specific coding concept or skill being tested, e.g. 'reading CSV files and computing column statistics with pandas', 'writing a bash script to recursively process files with pandoc', 'implementing UTF-16 endianness detection by analyzing byte order marks'\textgreater{}",\newline
~~"reasoning\_complexity": "\textless{}describe what level of coding reasoning this task requires, e.g. 'multi-step: requires parsing input format, building data structure, applying transformation, formatting output', 'single-step: direct library call with correct parameters'\textgreater{}",\newline
~~"error\_type": "\textless{}describe what went wrong in coding terms, e.g. 'wrong library function used for file traversal', 'failed to handle edge case in string encoding', 'produced pseudocode instead of executable code', 'incorrect regex pattern for parsing'\textgreater{}",\newline
~~"skills\_lacking": "\textless{}describe what coding knowledge or abilities the student is missing, e.g. 'the student needs practice with Python's os.walk for recursive directory traversal and subprocess for calling external tools'\textgreater{}",\newline
~~"what\_would\_help": "\textless{}describe what kind of coding training data would help, e.g. 'Python scripts that automate file format conversion using subprocess and os.path, with complete working examples showing error handling and edge cases'\textgreater{}",\newline
~~"summary": "\textless{}2-4 sentence explanation of what went wrong in coding terms --- what the code should have done, what the student produced instead, and what coding approach would have been correct\textgreater{}"\newline
\}
\end{promptbox}
\end{table}

\begin{table}[htbp]
\caption{\textbf{Reflection (policy) prompt.} System prompt used by the reflection model to turn aggregated error analyses into a weighted retrieval policy.}
\label{tab:prompt_reflection}
\begin{promptbox}
You are optimizing a data selection policy for training a student AI model.\newline
\newline
The policy is a list of (prompt, weight) pairs:\newline
~~- Each "prompt" describes what kind of training data to select from a large pool via embedding similarity search.\newline
~~- The "weight" controls what fraction of the training budget goes to that prompt.\newline
~~- Weights must sum to \textless{}= 1.0. The remainder is filled with randomly selected data.\newline
\newline
Based on the error analytics provided, propose a data selection policy. You should:\newline
~~- ADD prompts to cover the error patterns observed\newline
~~- ADJUST weights to reflect how frequently each pattern appears\newline
\newline
Guidelines:\newline
~~- Focus on what recurs across errors. A prompt can target a recurring domain (e.g. "contract law case analysis"), a recurring skill (e.g. "multi-step unit conversions"), or both if the cluster is tight enough (e.g. "thermodynamics with unit conversions between energy systems").\newline
~~- Each prompt should cover MULTIPLE related errors, not a single problem.\newline
~~- Aim for ${\sim}25$ prompts. Together they should cover all the major error clusters, not just the most frequent ones.\newline
~~- Look at the error examples to identify PATTERNS, not individual failures.\newline
~~- Give higher weights to patterns that appear more frequently in the errors.\newline
~~- Each prompt should describe training data in terms of TOPIC (e.g. "number theory", "graph algorithms"), SKILLS (e.g. "modular arithmetic, CRT"), and COMPLEXITY (e.g. "multi-step reasoning with case analysis"). Avoid vague or meta-skill prompts like "write correct code", "ensure proper formatting", or "verify final answers".\newline
\newline
Explain your reasoning, then provide the policy as a JSON array:\newline
[\newline
~~\{"prompt": "description of desired training data", "weight": 0.XX\},\newline
~~...\newline
]
\end{promptbox}
\end{table}

\section{Training Data Details}
\label{app:data_details}

We provide additional details on the data mixtures used across the training
stages. All corpora are drawn from publicly available open-source datasets.

\textbf{Pre-training.} The 7.1T-token pre-training corpus comprises web data, mathematics
data, code data, curated SFT-style data, and data from other domains: Nemotron-CC-v2~\citep{su2025nemotroncc,nemotronnano2} (web);
Nemotron-CC-Math-v1~\citep{mahabadi2025nemotronccmath}, MegaMath~\citep{zhou2025megamath},
and FineMath~\citep{lozhkov2024finemath} (mathematics); RefineCode~\citep{huang2025opencoder} and
Nemotron-Pretraining-Code-v1~\citep{nemotronnano2} (code); Nemotron-Pretraining-SFT-v1~\citep{nemotronnano2} (curated
pre-training SFT-style data); and the TxT360~\citep{txt360} subsets (other domains, including
\texttt{arxiv}, \texttt{dm\_maths}, \texttt{europarl}, \texttt{freelaw},
\texttt{hackernews}, \texttt{pg19}, \texttt{phil\_papers}, \texttt{pubmed},
\texttt{s2orc}, \texttt{stackexchange}, \texttt{ubuntu\_irc}, \texttt{uspto}, and
\texttt{wikipedia}).

\textbf{Mid-training variants.} The three mid-training variants share the Dolma~3
Dolmino 100B~\citep{olmo2025olmo} base mixture and differ only in a few STEM/reasoning subsets, as summarized
in Table~\ref{tab:midtrain_variants}. All other subsets are identical across the
variants.

\begin{table}[t]
\centering
\small
\caption{\textbf{Mid-training data-mixture variants.} Token counts for the
STEM/reasoning subsets that differ across the three mid-training variants; all
remaining Dolma~3 Dolmino 100B subsets are unchanged. \emph{Full pool} lists the
size of each subset in the full Dolma~3 Dolmino pool for reference.}
\setlength{\tabcolsep}{6pt}
\begin{tabular}{@{}llrrrr@{}}
\toprule
\textbf{Type} & \textbf{Source} & \textbf{v1} & \textbf{v2} & \textbf{v3} & \textbf{Full pool} \\
\midrule
Math (synth) & MegaMatt & 1.73B & 1.73B & 3.88B & 3.88B \\
Web pages & STEM-Heavy Crawl & 4.99B & 5.21B & 5.21B & 5.21B \\
Thinking (synth) & General Reasoning Mix & 1.87B & 1.87B & 2.48B & 2.48B \\
 & Math Meta-Reasoning & 381M & 381M & 1.05B & 1.05B \\
 & Code Meta-Reasoning & 459M & 459M & 1.27B & 1.27B \\
 & All other subsets & unchanged & unchanged & unchanged & --- \\
\midrule
\textbf{Total mix} & & 99.95B & 100.17B & 104.41B & --- \\
\bottomrule
\end{tabular}
\label{tab:midtrain_variants}
\end{table}

\textbf{Long-context Stage~2.} The Stage~2 mixture is a 37.32B-token curated blend
of math, code, and reasoning data drawn from the Dolma~3 Dolmino 100B mix, the full
Dolma~3 Dolmino pool~\citep{olmo2025olmo}, and Instella-GSM8K-synthetic~\citep{liu2025instella}.
Table~\ref{tab:longcontext_data_phase2} reports the mixture at the level of individual
source datasets, whereas Table~\ref{tab:longcontext_data} in the main text groups the
same 37.32B tokens by domain.

\begin{table}[t]
\centering
\small
\caption{\textbf{Long-context Stage~2 data mixture.} Token counts per subset for
the curated 37.32B-token math/code/reasoning mixture. \emph{Source} indicates
whether the subset is drawn from the Dolma~3 Dolmino 100B mix, the full Dolma~3
Dolmino pool, or Instella-GSM8K-synthetic.}
\setlength{\tabcolsep}{6pt}
\begin{tabular}{@{}lllr@{}}
\toprule
\textbf{Type} & \textbf{Dataset} & \textbf{Source} & \textbf{Tokens} \\
\midrule
\multirow{6}{*}{Math} & dolmino-math & 100B mix & 10.7B \\
 & cranemath & 100B mix & 5.62B \\
 & megamatt & Full pool & 3.88B \\
 & tinymath-mind & 100B mix & 898M \\
 & tinymath-pot & 100B mix & 241M \\
 & instella-gsm8k-synthetic & Instella-GSM8K & 329M \\
\midrule
Code & cranecode & 100B mix & 10.0B \\
\midrule
\multirow{4}{*}{Thinking} & general\_reasoning\_mix & Full pool & 2.48B \\
 & omr-rewrite-fullthoughts & 100B mix & 850M \\
 & math-meta-reasoning & Full pool & 1.05B \\
 & code-meta-reasoning & Full pool & 1.27B \\
\midrule
\textbf{Total mix} & & & 37.32B \\
\bottomrule
\end{tabular}
\label{tab:longcontext_data_phase2}
\end{table}

\textbf{SFT Phase~1.} The phase-1 SFT mixture combines a general
instruction-following dataset with three targeted skill slices spanning mathematics,
code, and science (Table~\ref{tab:sft_data}). In Phase~2, the model is annealed on the
feedback-driven curated 512K-example mixture described in
Appendix~\ref{app:sft_curation}, which is selected from a candidate pool comprising
Dolci-Think-SFT-7B~\citep{olmo2025olmo},
Nemotron-SFT-Competitive-Programming-v2~\citep{nemotronsftcp2}, and
Nemotron-Cascade-2~\citep{yang2026nemotroncascade2}.

\begin{table}[t]
\centering
\small
\caption{\textbf{SFT Phase~1 data mixture.} Record counts and the capability contributed by each
source. Mathematics records are reservoir-sampled from Nemotron-Cascade-2 with
reasoning prefixes stripped; all competitive-programming records that survive the
32K-token filter are retained; and science records are sampled from Nemotron-Cascade-2.}
\setlength{\tabcolsep}{6pt}
\begin{tabular}{@{}lp{0.30\linewidth}rp{0.30\linewidth}@{}}
\toprule
\textbf{Domain} & \textbf{Source} & \textbf{Records} & \textbf{Contribution} \\
\midrule
General & Dolci-Think-SFT-7B & $\sim$2.27M & General instruction-following base \\
Math & Nemotron-Cascade-2 (math) & 300K & Math; reservoir-sampled with the reasoning prefix stripped \\
Code & Nemotron-SFT-Competitive-Programming-v2 (Python) & $\sim$161K & Coding; all records surviving the 32K-token filter \\
Science & Nemotron-Cascade-2 (science) & $\sim$197K & Science reasoning and question answering \\
\bottomrule
\end{tabular}
\label{tab:sft_data}
\end{table}

\textbf{Post-training preference and RL data.} DPO uses the contrastive preference
pairs from Dolci-Think-DPO-7B~\citep{olmo2025olmo}. RL uses Dolci-Think-RL-7B: the IF-RL stage trains
the instruction-following expert on the \texttt{IF RLVR Mixture} (the
\texttt{IF\_multi\_constraints} slice), and the MOPD stage mixes the
\texttt{IF RLVR Mixture} with the general prompts at an IF fraction of $0.5$, with
each prompt domain-tagged so the domain router sends it to either the IF-RL teacher or
the frozen DPO anchor teacher.

\end{document}